\documentclass{article}

\usepackage{arxiv}

\usepackage[utf8]{inputenc} % allow utf-8 input
\usepackage[T1]{fontenc}    % use 8-bit T1 fonts
\usepackage{hyperref}       % hyperlinks
\usepackage{url}            % simple URL typesetting
\usepackage{booktabs}       % professional-quality tables
\usepackage{amsfonts}       % blackboard math symbols
\usepackage{nicefrac}       % compact symbols for 1/2, etc.
\usepackage{microtype}      % microtypography
\usepackage{lipsum}
\usepackage{graphicx}
\graphicspath{ {./images/} }

\usepackage{bm}
\usepackage{amsmath, amssymb, amsthm, enumitem}
\DeclareMathOperator{\E}{\mathbb{E}}
\usepackage{booktabs} % For professional looking tables
\usepackage{multirow}

\usepackage{floatrow}
\usepackage[font=large, labelfont={sf, sc, bf}, labelformat=simple]{subfig}% <-- changed
\usepackage{caption}
\title{Data-driven Emotional Body Language Generation for Social Robotics}

\author{
Mina Marmpena \\
University of Plymouth \\
Plymouth, UK \\
\texttt{mina.marmpena@itml.gr} \\
\And
Fernando Garcia \\
ANYbotics AG \\
Zurich, Switzerland \\
\texttt{fgarcia@anybotics.ch} \\
\And
Angelica Lim \\
Simon Fraser University \\
British Columbia, Canada \\
\texttt{angelica@sfu.ca} \\
\And
Nikolas Hemion \\
dSPACE GmbH \\
Paderborn, Germany \\
nhemion@dspace.de \\
\And
Thomas Wennekers \\
University of Plymouth \\
Plymouth, UK \\
\texttt{thomas.wennekers@plymouth.ac.uk} \\
}

\begin{document}
\maketitle
\begin{abstract}
In social robotics, endowing humanoid robots with the ability to generate bodily expressions of affect can improve human-robot interaction and collaboration, since humans attribute, and perhaps subconsciously anticipate, such traces to perceive an agent as engaging, trustworthy, and socially present. Robotic emotional body language needs to be believable, nuanced and relevant to the context. We implemented a deep learning data-driven framework that learns from a few hand-designed robotic bodily expressions and can generate numerous new ones of similar believability and lifelikeness. The framework uses the Conditional Variational Autoencoder model and a sampling approach based on the geometric properties of the model's latent space to condition the generative process on targeted levels of valence and arousal. The evaluation study found that the anthropomorphism and animacy of the generated expressions are not perceived differently from the hand-designed ones, and the emotional conditioning was adequately differentiable between most levels except the pairs of neutral-positive valence and low-medium arousal. Furthermore, an exploratory analysis of the results reveals a possible impact of the conditioning on the perceived dominance of the robot, as well as on the participants' attention.   \end{abstract}

\section{Introduction}
In the near future more robots are expected to be deployed in social environments such as schools, care homes and shops. Their ability to interact socially with humans is becoming increasingly important \cite{Breazeal} and as a consequence the field of social robotics is receiving increasing attention. Bartneck and Forlizzi \cite{Bartneck2004} maintain that a social robot must adhere to ``the behavioural norms expected by the people with whom the robot is intended to interact'', while Breazeal et al.\ \cite{Breazeal_social_2016} point out that social robots are expected to communicate with humans in an interpersonal manner, to engage them as partners, to collaborate or coordinate with them in order to accomplish positive outcomes. 
Among the characteristics of socially interactive robots identified by Fong et al.\ \cite{Fong}, we distinguish the ability to express and/or perceive emotions, establish/maintain social relationships, and exhibit personality and character. With our current work, we  aim to facilitate these aspects of a social robot by endowing humanoid robots with the capability to display believable, diverse and appropriate emotional body language (EBL).  

Body language is a subset of a broader group of nonverbal signals involved in human communication. Other such signals include facial expression or vocalizations. Facial expression is by far the most widely studied nonverbal signal of affect, with affective bodily expression lagging behind \cite{Karg2013}, although it constitutes a powerful communication channel among humans \cite{Kleinsmith, Witkower_Tracy, Coulson2004}. From a developmental point of view, body configurations are perceived from a very early age \cite{Gliga2005}, and they comprise the most primitive behavioural channel for humans to express emotion \cite{Fast1970}. Compared to facial expression, EBL can reveal more about the actual affective state \cite{ Argyle1975, Bull1977, Fast1970, Spiegel1974}, even when it is contradicted by facial expression and verbal communication \cite{Argyle1975}. Gestures and body postures evoke more trust \cite{Dael2012}, and they are less susceptible to social editing compared to facial expression \cite{Ekman1974}. Changes in a communicator's body orientation, posture configuration, proxemics, etc. can influence the overall likeability, interest, and openness between individuals \cite{Argyle1975, Bull1977, Mehrabian2007, Spiegel1974}. Bodily expression of emotion has also been found more effective than facial expression in discriminating between intense positive and negative emotions \cite{Aviezer2012} and in communicating emotion from a distance \cite{deGelder2010, Walk1988}.  

These insights on the effects of human EBL, taken together with the human tendency to attribute anthropomorphic traits to non-human agents \cite{Heider1944, Damiano2018, Koppensteiner2011}, support the hypothesis that the integration of robotic EBL in human-robot interaction (HRI) can enhance the user experience and make it more appealing and engaging \cite{Hortensius2018}. Evidence from studies in HRI show that bodily expressions of affect can draw human's attention \cite{Saad2019}, they are essential in naturalistic social interaction for robots that lack expressive faces (appearance-constrained) \cite{Bethel2008}, and can make the interaction more enjoyable \cite{Lee2006, Moshkina2005}. 

Believable robotic EBL needs to be lifelike, adjusted to the robot morphology, and granular to sustain attention in long-term interaction. Most importantly, it needs to be targeted, in the sense that the robot must be able to chose particular classes of emotions for different situations. In this work, we implemented a Conditional Variational Autoencoder (CVAE), a deep learning framework, to learn a latent, low-dimensional probabilistic representation of EBL animations for a Pepper robot. The model was trained with a small set of hand-designed expressions and its latent space can be sampled to generate numerous new animations of targeted valence and arousal. We present the extensive results of a user study designed to evaluate the interpretability of the generated animations, how they compare with designed animations in terms of anthropomorphism and animacy, and to what extent they are considered as emotional and they draw users' attention. 

The paper is organized as follows: Section 2 briefly discusses related work and Section 3 describes the theoretical aspects of the CVAE. In Section 4 we outline the methodologies we used for structuring the training set and sampling the CVAE, and we describe the design of the user study and the statistical methods for the data analysis. The results are presented in Section 5. 

\section{Related work}
Robotic EBL design has attracted much attention in recent years since for many robots, body motion is the main channel of non-linguistic expression. Nonetheless, comparing the proposed methodologies can be challenging due to differences in the design principles, the robot's level of embodiment, the modalities involved in the expressions, the type of the expressions (static postures or dynamic animations), and the evaluation methodologies. Another important aspect of variability is the emotion representation model adopted by the researchers both in the development and the evaluation phase of the synthetic EBL expressions. The most broadly used model is the categorical representation, which is based on the discrete basic emotions theories. According to it, there is a core set of distinct emotion categories, the basic emotions, each corresponding to distinct brain locations or networks, each manifesting with a specific feeling and neuro-physiological signature \cite{Reisenzein2014}. Proponents of the theory maintain that this multimodular emotion system is universal within the human species and perhaps also shared by other primates \cite{Ekman2011, Panksepp2011}. They also consider the basic emotions as prewired responses to different stimuli, an ``affect program'' written by the evolutionary process through natural selection, that can be however somewhat influenced epigenetically by learning. Arguably, the most frequently used emotion categories in the robotic EBL research is Ekman's list of six basic emotions: anger, disgust, fear, happiness, sadness and surprise \cite{Ekman1971}.

Another popular emotion representation in robotic EBL research is based on the dimensional models of emotion, in which affect states are represented by one or more continuous dimensions, each delimited by two polar states, such as pleased-displeased for the dimension of valence, activated-deactivated for the dimension of arousal, dominant-submissive for the dimension of control, approach-avoidance for the dimension of motivation. The most broadly adopted dimensional model is the \textit{circumplex model of emotion} \cite{Russell1999, Russell2003}, implemented as a two-dimensional space defined by the valence dimension ranging on a continuum from pleasure to displeasure on the horizontal axis, and the dimension of arousal ranging from activation to deactivation on the vertical axis. Russel and Barrett \cite{Russell1999} suggest that these two affective dimensions may be sufficient to capture the core affect, which becomes a full-blown emotion when it is assessed within a situational framework. This representation allows us to depict core affect as a data point on this 2D space, which is visually intuitive and readily informative, and makes clear how different emotions are not independent from each other, but instead, they are variations of the same two variables. Another popular dimensional model is the PAD emotional state model \cite{Mehrabian1974}, which uses an additional dimension, dominance, to support more nuanced emotions, and discern states like anger and fear which are both of low valence and high arousal (i.e., fear is presumably of low dominance, while anger is of high). 

In the following brief review, we summarize related work in robotic EBL synthesis under four main approaches: 1) Direct human imitation, 2) Feature-based design, 3) Creative design, and 4) Deep learning generative models. In many cases, researchers fuse elements from more than one approaches. The rest of this section discusses briefly each approach and provides some characteristic examples with a main focus on methods applied on humanoid robots and studies that include a user evaluation in terms of emotion recognition. For a broader systematic review on robotic animation techniques, the interested reader is directed to Schulz et al.\ \cite{Schulz2019}.

\subsection{Direct human imitation}
In the first approach, the robot joints are configured to match the human joints in a single posture or motion setting. To accomplish this, computer vision techniques, markers, or sensors are employed to track key joint positions in human body motion, and then map them to the robot joint space manually by observation \cite{Kleinsmith2009, Beck2012, Beck_towards_2010}, or by way of a transfer function \cite{Matsui2018}. Regarding the second method, although it has an advantage as an end-to-end process, constructing good transfer functions can be very challenging because of the extensive differences in the morphology and kinematics. To the best of our knowledge, there are no studies so far applying it specifically for robotic EBL synthesis. 

\subsection{Feature-based design}
A prominent approach in robotic EBL design is to use features extracted from human EBL. Such features can be found in studies coding postures or patterns of movement from recordings of actors performing emotional expressions \cite{Darwin1872, Meijer1989,Dael2012,Wallbott1998, Wasala, Gunes2007}, computer-generated mannequin figures \cite{Coulson2004}, or human motion capture data \cite{Kleinsmith_DB}. These studies provide lists of validated features, such as body orientation, the symmetry of limbs, joint positions, force, velocity, which can be employed for the design of robotic EBL expressions.  

Regarding the emotion representation in humanoid robots, most of the studies use the categorical model (e.g., Ekman's basic emotions) to design their robotic EBL expressions  \cite{Haring, Embgen, Erden2013, Tsiourti2017, McColl2014, Destephe2013}, but some adopted the dimensional representation in the evaluation \cite{Haring}. The robotic platforms used in these studies were Nao \cite{Haring, Erden2013}, Daryl \cite{Embgen}, Pepper and Hobbit \cite{Tsiourti2017}, Brian 2.0 \cite{McColl2014}, and WABIAN-2R \cite{DestepheROBIO2013}. In the evaluation, some of the studies displayed the designed expressions on a physical robot \cite{Haring, Embgen}, while others on a virtual robot, videos or images of the physical robot \cite{Erden2013, McColl2014, DestepheROBIO2013}. The evaluation results in these studies vary a lot, something that comes as no surprise considering the differences in the platforms and the design principles. Häring et al.\ \cite{Haring} found that all the body movements were rated in the right octant of the PAD model except \textit{sadness} which was rated with positive arousal. Embgen et al.\ \cite{Embgen} reported that participants were able to identify the emotions from the EBL and Erden \cite{Erden2013} that \textit{anger} was recognized with 45\%, \textit{happiness} with 72.5\%, and \textit{sadness} with 62.5\%. Tsiourti et al.\ \cite{Tsiourti2017} found that \textit{happiness} was recognized more accurately, while \textit{sadness} and \textit{surprise} were poorly recognized from body motion alone. McColl and Nezat\ \cite{McColl2014} found that \textit{sadness} and \textit{surprise} had the highest recognition rate ($>80\%$), while \textit{fear} and \textit{happiness} had the lowest recognition rate ($<30\%$). Destephe et al.\ \cite{DestepheROBIO2013} reported a high recognition rate for all the emotions (72.32\% average).

Another feature-based paradigm for robotic EBL design is inspired by the Laban Motion framework \cite{Laban}, based on kinesiology, anatomy, and psychological analysis of human motion. Often, researchers use only a subset of the framework, the Laban Effort System, which provides four motion parameters: space, weight, time, and flow. These parameters have been used to handcraft features and generate locomotion trajectories that express affect in non-humanoid robots with low degrees of freedom \cite{Knight2014,Sharma,Angel-Fernadez,Novikova,Takahashi2010}, with promising results in terms of readability by humans. For example, Sharma et al.\ \cite{Sharma} used the four parameters to configure a set of flying patterns for a Parrot AR.Drone, and they found in their dimensional evaluation that space and time were significant predictors of valence, while all four Laban parameters could predict arousal. Takahashi et al.\ \cite{Takahashi2010} designed expressions to convey the six basic emotions for a teddy bear robot using the Laban principles and they reported that although \textit{fear} and \textit{disgust} had a low recognition rate, the rest of the emotions were well recognised. The Laban Motion framework has also been used to design EBL for humanoids \cite{Masuda, Masuda2010b, Masuda2010d, Nomura2010, Nakata2001}. Masuda et al.\ \cite{Masuda} used Laban features (space, time, weight, inclination, height, area) to modify three basic movements for the humanoid robot HFR to express emotion. They found that \textit{sadness} had the highest modulating effect. 

Another interesting approach is to extract features from other human signals, such as voice, and encode them into robotic motion. The SIRE model \cite{Lim2014} was proposed for generating robotic EBL based on four voice features: speed, intensity, regularity and extent. The model was evaluated on a Nao robot using the dimensional PAD model, and the results showed successful recognition of \textit{happiness} and \textit{sadness}. 

Feature-based design can also be inspired by animal models instead of human EBL, an approach which can be potentially more effective with pet-like or animal-like robots \cite{Miklosi2012}. Lakatos et al.\ \cite{Lakatos2014} designed emotions of \textit{fear} and \textit{joy} for MogiRobi, a dog-like mechanical robot and they reported high recognition rates for \textit{joy}. Other interesting studies on emotion modelling for animal-like robots include the seal-robot Paro \cite{paro_1, paro_2}, the dog-robot AIBO \cite{Tamura2004}, the cat-robot NeCoRo \cite{necoro}. Although these studies have not directly evaluated emotion recognition and interpretability, they provide some indirect evaluation by testing the effects of using these emotionally expressive robots on robot-assisted therapy.

\subsection{Creative design}
Another influential approach in the design of robotic EBL emerges from applying the artistic principles and practices used in cartoon animation. The animator, conceives several expressive key postures for a given robot morphology, configures the robot like a puppet according to these postures to record the joints' positions, and then interpolates to derive intermediary positions so that the overall sequence appears continuous and smooth. This is the pose-to-pose animation technique. The approach is more robot-centric, in the sense that human EBL is not a direct prototype, and is often influenced by Disney's twelve basic principles of animation \cite{Disney} producing very lifelike expressions. In some studies, puppeteers \cite{Li, Thimmesch-Gill2017} or random participants \cite{Xu2015, Itoh2004} are enlisted to configure robot EBL postures or sequences of postures according to their subjective perception. 

Monceaux et al.\ \cite{Monceaux2009} describe their methodology of creating a big library of dynamic multimodal emotion expressions as an iterative process involving imitation from direct observation, and abstraction of key positions to adjust them to the Nao robot morphology. Ribeiro and Paiva \cite{Ribeiro2012} adapted Disney's twelve basic principles of animation \cite{Disney} for an EMYS head robot to express the six basic emotions, on three different intensities. Their user study evaluation found that \textit{anger}, \textit{sadness} and \textit{fear} were very well recognized. The principles of animation have also been used to design affect expression for non-humanoid robots. Yohanan and MacLean \cite{Yohanan2011} followed a pose-to-pose design for the Haptic Creature, an animal-like robot. The modalities involved ears, lungs and purr and the dimensional representation was adopted both in the design and the evaluation. The design was found effective in conveying arousal but ambiguous in the communication of valence.

\subsection{Deep learning approaches}
Lately, there have been some efforts to generate robotic EBL using deep learning. Suguitan et al.\ \cite{Suguitan2019} used human EBL for training three CycleGANs to generate animations expressing \textit{happiness}, \textit{sadness}, and \textit{anger} for a Blossom robot with four degrees-of-freedom. They reported that the intended emotions were fairly discernible. Marmpena et. al.\ \cite{Marmpena2019} trained a Variational Autoencoder (VAE) with a small set of robotic EBL animations created from professional animators for a Pepper robot to generate numerous new ones. The model was agnostic to the emotion class of the EBL, but after inspecting how the sampled trajectories from the learned 3D spherical latent space are decoded to the robot's joints' space, the authors propose that the latent space radius can be used as a sampling parameter to modulate the arousal content of the generated animations. Suguitan et al.\ \cite{Suguitan2020} also proposed a Variational Autoencoder, but they coupled it with a classifier to map the latent space motion representations to an emotion class. Their setup allows to modify a movement's emotion class by using latent space arithmetic. The user evaluation study showed that to some extent the modified movements are comparable to the original in terms of recognizability and legibility. 

\subsection{Contribution}
Previous research has been mainly focused on hand-coded methods which can be tedious and expensive, thus resulting in a limited number of expressions, characterised by less granularity in their expressivity. Consequently, the expressions might appear repetitive and predictable which might be disengaging in long-term human-robot interaction. Furthermore, many of the previous studies used human EBL as a prototype, but scaling down the human motion range or excluding joints to match the robot's simpler morphology might end up loosing information, or preserving redundant or trivial information with a negative impact on the lifelikeness of the robot character. 

Our current work proposes a data-driven deep learning methodology for the automatic generation of numerous new robotic EBL animations, which appear smooth, natural, and exhibit increased granularity. Instead of human EBL, the prototype used is based on creative design which is directly adapted to the robot morphology and kinematics to ensure the illusion of life effect. Importantly, the proposed methodology seeks to generate animations of specific emotion class, so that the robot can select appropriate responses to given signals. Our work also seeks to leverage the dimensional representation of emotion, and more specifically the circumplex model of core affect \cite{Russell1999, Russell2003} with the dimensions of valence and arousal. We use the representation as an integral component of the generative process.

Concretely, we build on the methodology proposed in \cite{Marmpena2019}, where a VAE model was trained with a small set of hand-designed motion animations which were created with direct adaptation to the Pepper robot morphology and motion capabilities. From the same work, we adopt the proposed method to modulate arousal by exploiting the geometry of the model's latent space. For valence, we expand the VAE model to a conditional VAE (CVAE) which takes as input a label of valence as a continuous scalar value, and learns to adapt the generated animations based on it. Furthermore, to improve the expressiveness of the generated animations, we augmented the motion training set with sequences of eye LEDs' colourful patterns.

\section{Conditional Variational Autoencoders}\label{s_cvae}
The Variational Autoencoder (VAE) framework \cite{Kingma2013, Rezende2014, Kingma2019} can be used to learn a posterior probability distribution that represents the unknown underlying process that generates the observed data. Subsequently, this distribution can be sampled to generate purely novel content which is similar to the original. The VAE is composed of an encoder and a decoder, which can be implemented with any neural network architecture. During training, the vectorized input $\bm{x}$ is passed through the encoder which compresses it into a lower-dimension stochastic representation $\bm{z}$, which retains all the sufficient information for a faithful  reconstruction of $\bm{x}$. The distribution of all $\bm{z}$ encodings defines a latent space, i.e., a continuous and structured manifold in which similar datapoints are close to each other. Then, the encoding $\bm{z}$ is passed though the decoder which uses the information compressed in it to reconstruct the input. 

In probabilistic graphical model formulation, the encoder is an inference model $q_\phi(\bm{z} \mid \bm{x})$ which approximates the true but intractable posterior. It infers the parameters of a predefined distribution, that is $\bm{\mu}$ and $\bm{\sigma}$ in the case of a multivariate Gaussian with a diagonal covariance matrix, which is the usual choice for real-valued output. Subsequently, the parameters are used to sample a latent datapoint $\bm{z}$, which is then passed through the decoder, a generative model $p_\theta(\bm{x} \mid \bm{z})$. Both $\phi$ and $\theta$, the parameters of the two models, are learned jointly during training. This is accomplished by minimizing the following loss function: 

\begin{equation}
\mathcal{L}_{\theta, \phi}(\bm{x}) = 
\underbrace{-\E_{\bm{z} \sim q_\phi(\bm{z} \mid \bm{x})}[\log{p_\theta(\bm{x} \mid \bm{z})}]}_{reconstruction~error} + 
\underbrace{D_{KL}[q_\phi(\bm{z} \mid \bm{x}) \parallel p_\theta(\bm{z})]}_{regularization~term},
\label{eq:vae_loss_function}
\end{equation}

In Eq. \ref{eq:vae_loss_function}, the reconstruction error is the negative expectation of the log-likelihood taken over the latent parameters $\bm{z}$. It encourages the $\bm{z}$ values that contribute to the faithful reconstruction of $\bm{x}$ by the generative model. The regularization term (also called variational loss) is the Kullback-Leibler divergence which estimates the disparity between the latent distribution and a prior which is usually set to be a centered isotropic multivariate Gaussian. It incentivizes the inference model to learn $\bm{z}$ values that are close to the prior distribution, and this way it applies a constraint on the latent space structure, resulting in a continuous manifold from which we can sample and generate new realistic samples. 

The VAE is a powerful generative model, but in its standard form, there is no control on the class of the generated output. Controlling the generative process so that the decoded output matches a given class can be very important in certain applications. For example, in the present work, we want to be able to generate animations of targeted valence. The Conditional Variational Autoencoder (CVAE) \cite{Kihyuk2015}, an extension of the VAE framework, does exactly what we need by learning to make a discriminative prediction additionally to reconstructing the input. This is accomplished by using an auxiliary variable $\bm{c}$, essentially a continuous or discrete label, to condition the encoding and decoding processed. The loss function of the CVAE modifies Eq. \ref{eq:vae_loss_function} as follows: 

\begin{equation}
\mathcal{L}_{\theta, \phi}(\bm{x},\bm{c}) = 
-\E_{z \sim q_\phi(\bm{z}\mid\bm{x},\bm{c})}[\log{p_\theta(\bm{x}\mid\bm{z}, \bm{c})}] + 
D_{KL}[q_\phi(\bm{z} \mid \bm{x}, \bm{c}) \parallel p_\theta(\bm{z})].
\label{eq:cvae_loss_function}
\end{equation}

\begin{figure}[htbp!] 
\centering    
\includegraphics[width=0.6\textwidth]{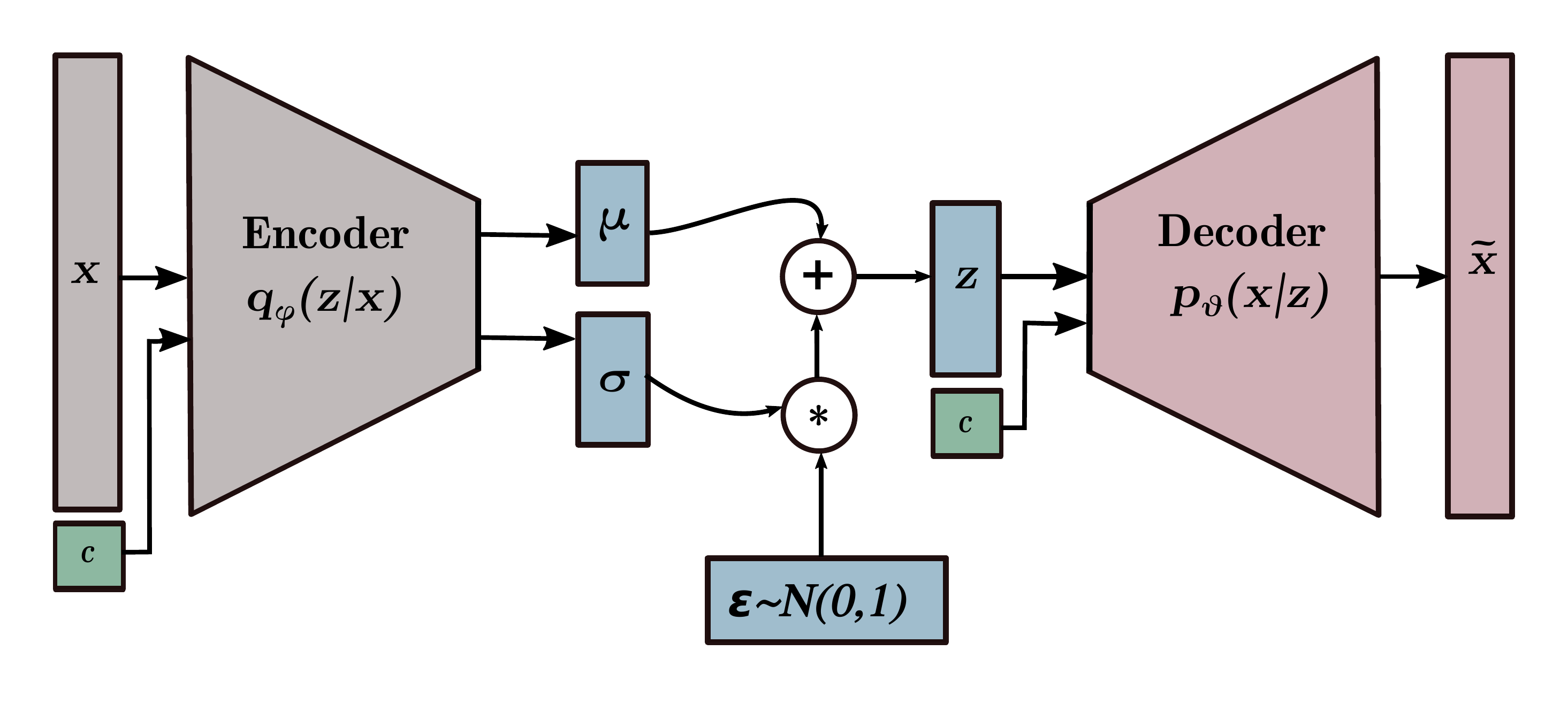}
\caption{The structure and computational flow of the CVAE. Compared to the standard VAE, there is an additional input $\bm{c}$ (green block) that holds a label or an attribute on which we want to condition the generative process. }
\label{fig:CVAE}
\end{figure}

Fig.~\ref{fig:CVAE}, illustrates the structure and computational flow of the CVAE. During training, at each step, we concatenate the input $\bm{x}$ and  the label $\bm{c}$, and we feed them as one vector to the encoder. The output of the encoder is reparameterized with some random noise $\bm{\epsilon}$ to produce $\bm{z}$, which is concatenated with $\bm{c}$ again and fed to the decoder, which outputs the reconstructed $\bm{x}$. After training, generating new output with label $\bm{c}$ is accomplished by sampling a $\bm{z}$ from the multivariate Gaussian prior, concatenation with $\bm{c}$ and decoding.

\section{Methods and materials}
This section presents the training dataset and the implementation of the CVAE framework for robotic EBL generation, as well as the design of a user evaluation study and the statistical methods applied for the analysis of the collected data. 

\subsection{Training dataset}
\begin{figure}[htb]
    \centering
    \includegraphics[width=0.8\textwidth, trim={20 100 60 30},clip]{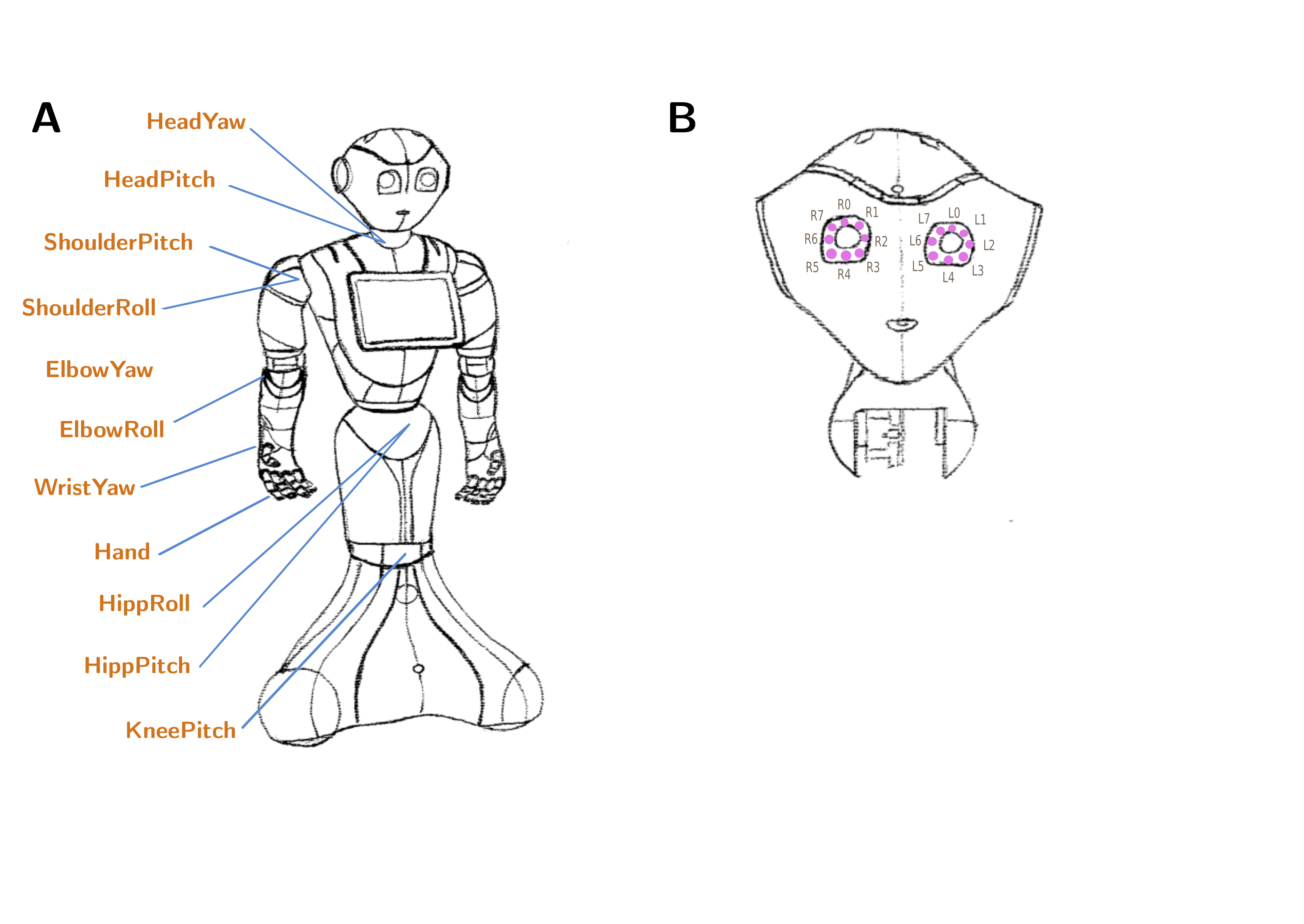}
\caption{A) A Pepper robot has 17 joints that control the body configuration. The drawing shows the StandInit posture. B) The eye LEDs of Pepper. The robot has 8 RGB LEDs around each eye.}
    \label{fig:pepper}
    \end{figure}

The dataset we used contains 36 robotic animations designed with the pose-to-pose method by professional animators for the Pepper robot. Initially, each animation consisted of three different modalities: motion, eye LEDs colour patterns, and non-linguistic sounds. Motion is represented by a series of keyframes, with each keyframe describing a posture defined by 17 values, i.e., the angles of the robot's joints in radians (Fig. \ref{fig:pepper}A). Each keyframe also has a timestamp indicating after how many frames (starting from 0) the posture should be executed. At runtime, the intermediary frame postures are obtained with cubic Bezier interpolation applied locally at each joint. The eye LEDs sequences are somewhat different in that their values can be set at any frame for a specified interval and they remain static until the interval is over. Pepper has two eyes with 8 LEDs each (Fig. \ref{fig:pepper}B). Each LED is described by 3 values representing an RGB colour. Thus, in total, there are 48 LED values within a range of $[0,1]$. For the current study, we excluded the audio modality. 

The particular set of animations has been previously annotated with valence and arousal ratings in a user study with 20 participants \cite{Marmpena2018}. In the present study, we only use the valence ratings as labels to condition the CVAE as described in Subsection \ref{s_cvae}. Each animation is annotated with a single real-valued label of valence in $[0,1]$, where 0 indicates very negative valence and 1 very positive. For arousal, we used a different method of conditioning, based on the topology of the latent space (more details are provided in the next subsection). 

% Recording
Our goal for the CVAE training was to use postures as input at each timestep to allow the network to learn a continuous and structured manifold of postures, which can be interpolated to generate smooth sequences of motion. The initial keyframe representation was unsuitable for that, due to the abrupt changes between subsequent keyframe postures, and also the small size of the dataset. Furthermore, since the eye LEDs changes did not have the same onset with the keyframe postures, a dataset with the initial representation would be very sparse. To address these issues, we had the animations executed by the real robot and we recorded the angles of the joints and the eye LEDs values at a constant sampling rate of 25 frames per second, thus capturing both keyframes and intermediary frames. Before the recording, we made a small modification in the keyframe representation by adding an extra keyframe describing the StandInit posture (see posture in Fig.~\ref{fig:pepper}) at the beginning and the end of each animation (whenever it was missing), to obtain a clear onset and offset of an animation. In total, the initial keyframe representation dataset contained 697 keyframes, and with the recording we obtained 5074 frames. At this stage, each frame was a vector of 65 values: 17 values for the angles of the joints and 48 for the eye LEDs state.  

The next issue to address was related to the prevalence of possible body orientations. For example, if many postures assume a right-side orientation, the training could be heavily biased and the network might not learn enough postures with left side orientation. We made the assumption that the interpretation of the bodily expression of emotion is independent of left or right side orientation with respect to the body's vertical axis. This assumption allowed us to augment the dataset by having every posture mirrored from left to right or the opposite. We avoided any mirroring with respect to the horizontal axis of the body, e.g., changing upward motion to downward or vice versa, since such bias could affect the emotion content interpretation. More specifically, to create a mirrored posture we swapped the values of the entire chain of arm joints between the left and right arm, i.e shoulder pitch and roll, elbow roll and yaw, wrist yaw, and hand. For the roll and yaw joints of the arm, we had to additionally invert the signs of their values because the ranges of these joints had inverse signs between left and right. Furthermore, we also mirrored the head yaw and hip roll joints since these also move right and left with respect to the vertical axis of the body. For this mirroring, only the signs needed to be inverted. This augmentation doubled the dataset to 10148 frames. 

% Valence attr
Regarding the valence attribute, we repeated the valence rating related to each animation for all the frames belonging to it. Therefore, each training example consists of 66 elements: 17 for the joints' angles describing the posture, 48 for the eye LEDs state, and one value indicating the valence rating of the animation to which the frame belongs.

% Normalization
The last step in the preprocessing pipeline was the data normalization. The ranges of plausible joint values differ among the joints, thus we rescaled all the values in [0, 1] with Min-Max Normalization. This was not necessary for the LEDs or the valence label since their range was already in $[0,1]$. Since the implemented CVAE is agnostic to sequential dynamics of the dataset, we shuffled the dataset and used 80\% for the training, and the rest for the validation.

\subsection{CVAE network implementation}
The CVAE framework for robotic EBL generation was implemented with Multilayer Perceptrons (MLPs). The decoder MLP has three dense layers with 128, 512, and 128 units respectively, as well as an additional dense layer after the output, serving as a reconstruction layer, with 66 units to match the input dimensions. The capacity of the encoder was increased by one extra fully-connected hidden layer with 512 units. This was decided after several empirical tests that showed that this configuration was more efficient to avoid posterior collapse\footnote{Posterior collapse is a well documented problem arising in VAE training. It appears when the the model instead of learning meaningful latent features from the input, learns to ignore the latent variables resulting in a posterior distribution that collapses to the prior (usually selected to be an uninformative prior, e.g., a unit Gaussian). More concretely, posterior collapse indicates that the training is trapped into a trivial local optimum where $q_\phi(\bm{z}\mid\bm{x}) \simeq p(\bm{z})$ for all $\bm{x}$. Empirically, the problem can be detected during training when the variational loss $D_{KL}[q_\phi(\bm{z} \mid \bm{x}) \parallel p_\theta(\bm{z})]$ becomes almost zero.}. After each hidden layer, we used a layer of rectified linear activation functions (ReLU), while the output layers were followed by linear activations. The reconstruction layer of the decoder uses a sigmoid function to rescale the output in the range [0,1]. After each hidden layer, we used a dropout layer, which randomly sets half of the units to zero \cite{hinton2012}. This was not necessary in terms of regularization, but it improved the speed of the training.

For the latent layer, we used a strong bottleneck with only 3 dimensions. More dimensions did not decrease the validation error significantly, while less dimensions increased it. As we discussed in Section \ref{s_cvae}, the latent layer infers the parameters of a multivariate Gaussian distribution (a 3D Gaussian in our case), so it consists of two sub-layers of 3 units each, one for the latent mean $\mu$ and another one for the latent standard deviation $\sigma$ used in the diagonal covariance matrix. In terms of the loss function implementation, the reconstruction error is equivalent to the Mean Squared Error (MSE) between the input and the reconstructed output when we assume a Gaussian model with real-valued output for the decoder. The regularization term can be computed analytically when we assume Gaussian distributions for both the prior and the approximate posterior \cite{Kingma2013}. Furthermore, the regularization term was weighted with a $\beta$ coefficient \cite{higgins2017, higgins2018}. In general, the larger this hyperparameter is, the stronger the pressure on the approximate posterior $q_{\phi}(\boldsymbol{z}\mid\boldsymbol{x})$ to match the prior $p_\theta(\boldsymbol{z})$. However, a large $\beta$ might also have a cost on the reconstruction fidelity. After several empirical tests, we tuned the hyperparameter to $\beta=0.001$. This value was found to allow a low reconstruction error and also a well structured latent space. 

The network weights were initialized with a Xavier uniform initializer \cite{glorot2010}, i.e., the initial weight of each unit was sampled from a uniform distribution with limits normalized by the sum of its input and output units. For the optimization, we used an Adam optimizer \cite{kingma2014} with a learning rate set to 1e-4. Finally, we trained for 250 epochs. 

\subsection{Conditional sampling of the latent space} \label{s_sampling}
The simplest way to generate a new animation with the trained CVAE is to randomly sample the prior distribution (centered isotropic 3D Gaussian) as a proxy for the learned latent space, interpolate the samples, concatenate the target valence label at each datapoint, and finally decode them into the joint space. However, a more systematic sampling method proposed in previous work \cite{Marmpena2019} uses concentric spherical grids projected conceptually on the latent distribution and samples latent trajectories along their longitudes. This work showed that trajectories sampled from the inner spheres, closer to the core (which is also the center of the latent zero-centered distribution), decode into animations with lower amplitude and variance compared to animations decoded from trajectories sampled from the outer spheres. Based on this finding, it was suggested that the radius of the projected spheres can be used as a topological feature that modulates the arousal content of a generated animation. Therefore, to generate low arousal animations, we use a small radius as standard deviation to sample the prior distribution, and when we want to increase the arousal we increase the radius. 

\begin{figure}[ht!] 
\centering    
\includegraphics[width=0.5\textwidth]{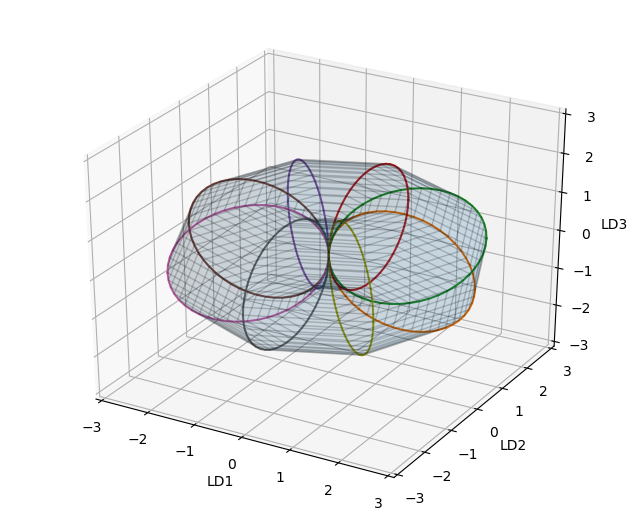}
\caption{An example of a horn torus grid used as a template for the systematic sampling of latent trajectories. In this example, the radius is equal to 3, and the axis is in parallel with the latent dimension LD3. All torus grids have their center in the core of the latent space where $LD1=LD2=LD3=0$. The latent trajectories are sampled along the colored longitude lines.}
\label{fig:torus_grid}
\end{figure}

In the current work, we test this hypothesis about the arousal conditioning but we modify the conceptual tool of the spherical grid into a torus. A shortcoming of the spheres was that the decoded animations appeared to begin and end abruptly, and always in the same robotic posture. This is due to the fact that all the longitudes of a sphere end at the same point, the pole of the sphere. Especially in larger spheres, this point is far away from the origin where all three latent dimensions (LD1, LD2, LD3) are equal to zero, i.e., the encoding that decodes into the StandInit position, the most neutral and symmetric posture. By changing the spherical grid topology into a torus grid, and more specifically, a \textit{horn torus} (see Fig.~\ref{fig:torus_grid}), which is a torus without a hole, with all the circles forming its tube touching each other at the centre of the 3D latent space where $LD1=LD2=LD3=0$, we get a clear advantage. Although all the longitudinal trajectories begin and end at the same point, and unlike the poles of the sphere, this point is now located in the core of the latent space, and thus, it decodes into the StandInit posture. This way, we obtain a clear onset and offset of the animation and we still keep the radius feature with which we will modulate the arousal dimension of the generated animations.   

% \subsubsection{Arousal conditioning}
To condition the generated animations on the arousal as hypothesized to be modulated by the radius of the latent space, we defined three levels of arousal (low, medium, and high), which were obtained by sampling the 3D latent space with a radius of 3, 4, and 5 respectively. More concretely, we sampled 8 circular latent trajectories from torus grids rotated in three ways (so that the central axis is parallel to one of three latent dimensions), and radiuses of 3, 4 and 5. For example, Fig.~\ref{fig:torus_grid} illustrates 8 circular trajectories from a torus grid of radius equal to 3, and its central axis in parallel with LD3. Note that in this approach we define radius as the distance from the centre of the latent space to the outer point of the torus surface as if the whole torus is enclosed in a sphere with the same centre. Each latent circle was defined by 20 points. Subsequently, the 20 points of each latent trajectory were interpolated using B-spline interpolation of 15, 20, 25 steps per segment for radius 3, 4, and 5 respectively. We gradually increased the interpolation steps as the radius increased to keep similar distances between the points of each interpolant. In total, we derived 72 latent trajectories (8 longitudes * 3 radius values * 3 latent dimensions) with 20 points each.  

% \subsubsection{Valence conditioning}
Subsequently, to condition the generative process with the valence label, we concatenated each of the 72 latent interpolants with 3 different values representing levels of valence: 0 for negative, 0.5 for neutral, and 1 for positive. Thus we obtain 216 latent trajectories which were passed through the decoder of the CVAE to get the final library of 216 generated animations. Essentially, the valence conditioning is achieved via the CVAE model structure, while the arousal conditioning is accomplished explicitly based on the topological features of the spherical latent space. 

\subsection{Evaluation study}
This section presents the design of a user study aimed to evaluate the interpretability of the CVAE generated animations with respect to the valence and arousal conditioning. The statistical methods applied for the analysis of the collected data are also presented.   

\subsubsection{Participant's profile}
A total of 20 volunteers were recruited for this study (9 female and 11 male). The mean age was 26.4 years ($SD = 5.36$, $min=21$, $max=44$). All participants were employees of SoftBank Robotics Europe, in various professional roles, e.g., engineers, administrative staff, marketing, etc. They self-reported their experience with the robot's animation capabilities on a scale ranging from 0 (no experience at all) to 10 (extremely familiar). The mean self-reported experience was equal to 6.35 ($SD = 1.81$, $min=3$, $max=9$). The experiment was carried out at the company's premises in Paris, France. The experimental protocol was granted ethical approval by the University of Plymouth, Faculty of  Science and Engineering Research Ethics Committee.

\subsubsection{General setup}
The experimenter briefly explained that the aim of this experiment was to evaluate a set of body language animations displayed by Pepper. This would require to observe the robot located in front of the participant, approximately 2.5 meters away, and reply to the questions appearing on a screen right in front of the participant. The session was split into three parts, all of which were completed by each participant. In general, when deciding on the number of trials, as well as the length of questionnaires, our main concern was to keep participant's workload as low as possible, so that they could retain their concentration and motivation. The interface was implemented as a website running on a local server and it was projected on the screen. The participant would click on a play button to give the command for the robot to execute an animation, and afterwards, at her own pace, she would insert her responses. 

\subsubsection{Stimulus, interface, and questionnaires}\label{s_stimulus}
The stimuli used for this experiment were animations displayed on a physical Pepper robot. More specifically, we had two kinds of animations: \textit{designed} animations, designed by professional animators with the pose-to-pose method, and \textit{generated} animations, generated with the CVAE and the sampling method we described in section \ref{s_sampling}. 

\textbf{Part A}: This part aimed to compare the \textit{designed} animations to the \textit{generated} ones in terms of the anthropomorphism and animacy degree attributed to the robot. It consisted of two trials, in which participants watched the robot executing 9 \textit{designed}, and 9 \textit{generated} animations. In both groups, the animations were displayed continuously without a pause. The animations were randomly selected in each group from the respective broader libraries of \textit{designed} and \textit{generated} animations. The two trials appeared in randomized order. After each trial, participants had to fill in a questionnaire comprised of two scales, Anthropomorphism and Animacy, selected from the Godspeed Questionnaire Series (GQS) \cite{Bartneck2009}. Anthropomorphism is a measure of how humanlike the robot appears to be. The scale comprises of five 5-point semantic differentials (Fake/Natural, Machinelike/Humanlike, Unconscious/Conscious, Artificial/Lifelike, and Moving rigidly/Moving elegantly). Animacy is a measure of how lifelike the robot appears to be. The scale comprises of six 5-point semantic differentials (Dead/Alive, Stagnant/Lively, Mechanical/Organic, Artificial/Lifelike, Inert/Interactive, Apathetic/Responsive). 

\begin{figure}
  \centering
    \fbox{\includegraphics[width=0.6\textwidth]{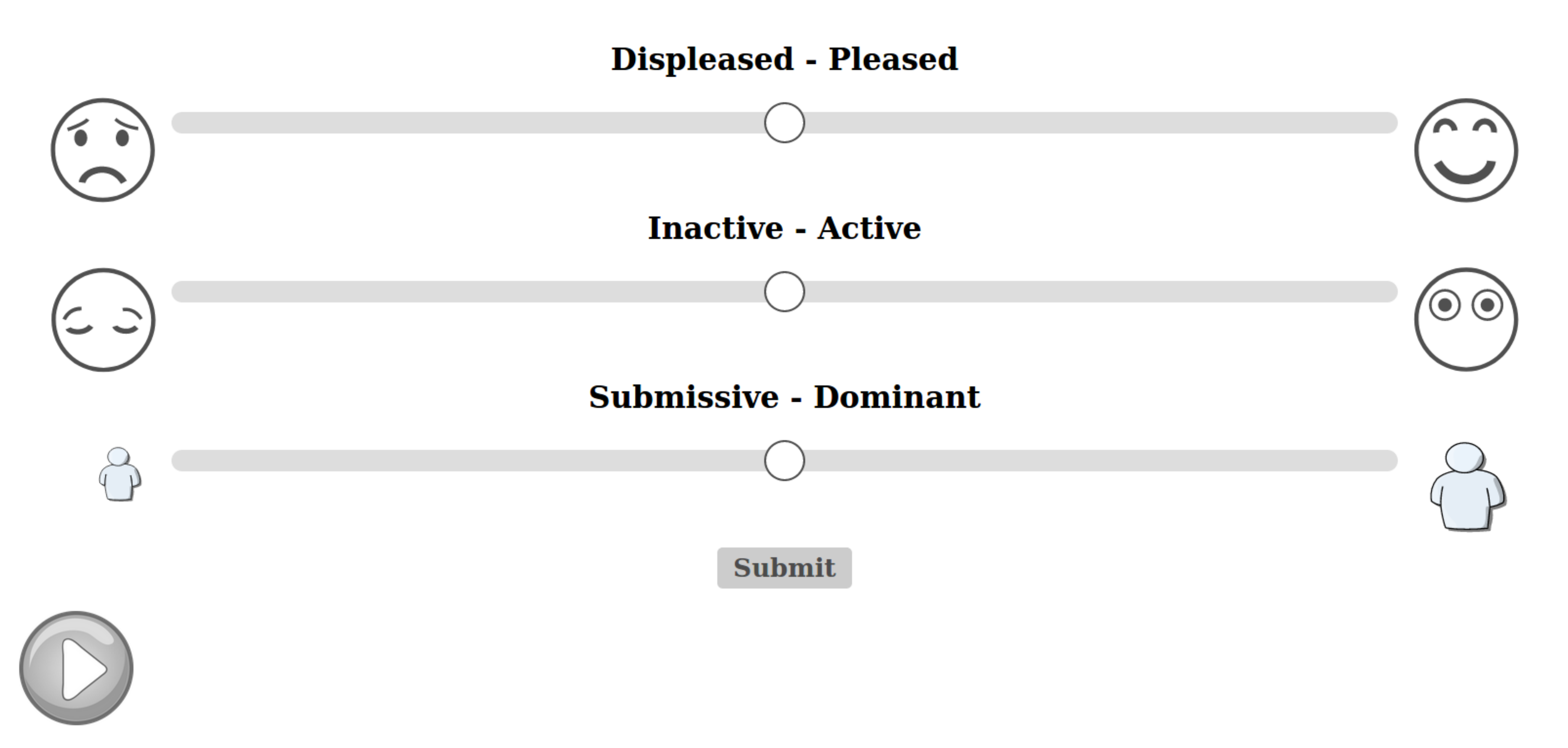}}
    \caption{The view we used to collect valence, arousal and dominance ratings. The participant clicks on the play button and watches the real robot executing a generated animation. Subsequently, the user clicks on the three sliders to enter the perceived valence, arousal and dominance scores. The concept is adopted from the Affective Slider \cite{Betella2016}.}
    \label{fig:sliders_CVAE}
\end{figure}

\textbf{Part B}: This part involved only \textit{generated} animations, which were evaluated on three dimensions of emotion: valence, arousal and dominance. It consisted of 18 randomized trials. In each trial, a single \textit{generated} animation was displayed on the robot when the participant clicked on a digital play button. After the animation was over, the participant would enter the dimensional ratings on three different sliders as depicted in Fig.~\ref{fig:sliders_CVAE}. The sliders were titled with the two extremes of each measured emotion dimension: for valence the title is \textit{Displeased - Pleased}, for arousal \textit{Inactive - Active}, and for dominance \textit{Submissive - Dominant}. We selected this verbal description as more intuitive. The interface uses visual aids to illustrate the extremes in a more intuitive way. The whole idea of this view is based on the Affective Slider \cite{Betella2016}, a digital scale for the self-assessment of emotion, which we modified by adding the extra dimension of dominance (\textit{Submissive - Dominant}) and also by making small changes to the original icons for valence and arousal to avoid misreadings between valence and arousal. The participants were allowed to click the play button and watch the animation for a second time if needed. After watching the animation they could click on the sliders to enter their ratings. Each slider was constructed with a range in $[0,1]$ and 100 points of discretization. However, the numerical representation was hidden. 

After submitting the ratings, a second view appeared in the same trial. It contained two 5-point Likert scales and the play button. The two declarations on the Likert scales were ``The robot's behaviour draws my attention'' and ``The robot's expression was emotional''. The responses for each of declaration were: Strongly disagree, Disagree, Neither agree nor disagree, Agree, Strongly agree. The participant could watch the animation once more if it was necessary to recall it in the context of the new set of questions. The first Likert declaration is adapted from \cite{Gomes2013}, a study that proposes several dimensions and related metrics to evaluate the believability of an artificial agent. The authors of this study, propose a Likert scale with the template phrase, ``<X>'s behaviour draws my attention.'' as a quantifiable metric of the \textit{visual impact} dimension of believability. The second Likert scale was constructed by us as a manipulation check, to obtain a secondary, more direct and general assessment on whether participants perceived a \textit{generated} animation as an emotional expression. The 18 \textit{generated} animations used in Part B were selected randomly from the broader set of generated animations, obtained as we described in Section \ref{s_sampling}. We aimed to have 2 animations for each combination of valence and arousal levels. The task was solicited with the question `How does the robot feel?'. A practice period of 3 trials preceded it so that the participants could get accustomed to the use of the sliders and the emotion dimension concepts.

\textbf{Part C}: This part was identical to Part A in terms of structure but used different \textit{designed} and \textit{generated} animations to minimize the chances of confounds due to the choice of animations. Furthermore, the \textit{generated} animations used in Part A and C were excluded. Part C, was added as a pretest-post-test design to examine if the results of the comparison between designed and generated samples (on anthropomorphism and animacy) persist or change after the participants become more familiar with the robot range of motion.  

\subsection{Statistical analysis}
The statistical analysis of the collected data examines: 1) How valence and arousal conditioning of the generative process influences participants in their evaluation of the generated animations' valence, arousal and dominance, 2) How the participants compare the generated vs the hand-designed animations in terms of anthropomorphism and animacy, and 3) How valence and arousal conditioning of the generative process influences the participants' attention and their belief that the animations express emotions. 

\subsubsection{Valence, arousal and dominance ratings}
This analysis aims to explore how valence and arousal conditioning of the generated animations affected the ratings of valence, arousal and dominance assigned by the participants. Our two independent variables are the following: 1) \textit{v\_cond} with levels \textit{negative}, \textit{neutral}, and \textit{positive} refers to the valence conditioning of the CVAE generative process using an attribute $c$ equal to 0, 0.5, and 1 respectively, and 2) \textit{a\_cond} with levels \textit{low}, \textit{medium}, and \textit{high} represents the arousal sampling of the CVAE's latent space using a radius $r$ equal to 3, 4, and 5 respectively.

There are three dependent variables, \textit{valence}, \textit{arousal} and \textit{dominance}, which are based on the corresponding slider ratings given by the participants. Since we had multiple responses from a participant within the levels of the independent variables (e.g., a participant rated 6 animations with \textit{v\_cond} = \textit{neutral}), we aggregated the ratings for each participant within each level of the two independent variables \textit{v\_cond} and \textit{a\_cond}. The aggregation was based on the mean within each level.

The main research question is whether the valence and arousal ratings given by the participants are affected by the valence and arousal conditioning we used to generate the animations. Specifically, we had two hypotheses:

\begin{itemize}
    \item {\textbf H1}: Valence ratings for animations generated with {\itshape v\_cond = negative} will be lower than those of animations generated with {\itshape v\_cond = neutral} or {\itshape v\_cond = positive}. Valence ratings for animations generated with {\itshape v\_cond = neutral} will be lower than those of animations generated with {\itshape v\_cond = positive}.
    \item {\textbf H2}: Arousal ratings for animations generated with {\itshape a\_cond = low} will be lower than those of animations generated with {\itshape a\_cond = medium} or {\itshape a\_cond = high}. Arousal ratings for animations generated with {\itshape a\_cond = medium} will be lower than those of animations generated with {\itshape a\_cond = high}.
\end{itemize}

We also conducted an exploratory analysis to examine whether \textit{v\_cond} had an effect on the arousal or dominance ratings, and similarly, if \textit{a\_cond} had an effect on valence or dominance ratings. Since the experimental design was within-subjects, i.e., each participant evaluated animations in each level of the independent variables, we used repeated-measures one-way analysis of variance (ANOVA). The test was decided to be one-way (with a single independent variable at a time) because the data aggregation applied on the levels of each independent variable could not be applied concurrently for both independent variables without inflating the number of appearances of each animation and inserting multiple missing values. In total, we applied 6 such tests and we conducted post hoc comparisons between the levels of the independent variables with paired samples t-tests, only for those relationships that were found significant. The \textit{p} values were adjusted with Bonferroni correction for multiple comparisons error. Before proceeding with the tests, we checked whether the necessary ANOVA assumptions hold. We checked for outliers using boxplot methods. The normality assumption was tested with the Shapiro-Wilk test and visual inspection of the QQ plots within each level of the independent variables. The assumption of sphericity was checked with Mauchly’s test and Greenhouse-Geisser sphericity correction was applied when the assumption was violated. 

\subsubsection{Comparison of designed and generated animations}
This analysis aims to examine the comparison between {generated} and \textit{designed} animations in terms of anthropomorphism and animacy. It also examines for possible effects arising from the pretest-posttest experimental design (Part A vs Part C as described in \ref{s_stimulus}) to determine if the attribution of anthropomorphism and animacy is altered after the participants become more familiar with the robot's body language. The main hypothesis for this analysis is the following:

\begin{itemize}
    \item {\textbf H3}: The attribution of anthropomorphism and animacy to the robot is not significantly different between the \textit{designed} and the \textit{generated} animations, either in the pretest trials (Part A) or the posttest trials (Part C).
\end{itemize}

Initially, we computed Cronbach's alpha to estimate the internal consistency of the responses as a measure of reliability. An acceptable reliability would permit to collapse the items of each scale to a single response per participant and per scale using a measure of central tendency. The decision was based on a threshold $\alpha > 0.7$ which is widely considered as an indication of acceptable internal consistency. Given that result, we proceeded with taking the median of the items of each scale and each participant, i.e., one score for Anthropomorphism and another one for Animacy for each participant. Since Likert scales provide ordinal data, the mean is not considered a valid parameter to aggregate their items, and similarly, parametric tests cannot be applied due to the normality assumption, thus we used ordered logistic regression \cite{Liddell2018} to model the effects. For pairwise differences of the estimated marginal medians, the \textit{p} values of the contrasts were corrected with Tukey's multiple comparisons adjustment.  

We fitted 3 models for each scale. The first one aimed to predict the Likert scores using the group of the animation set (designed\_pre, generated\_pre, designed\_post, generated\_post) as a predictor, and to examine if there are differences between the \textit{generated} and the \textit{designed} animations, either in the pretest phase or the posttest, according to \textbf{H3}. For the second model, we used as a predictor only two groups (pre, post), each containing both designed and generated animations, in order to explore if there are pretest-posttest differences in general. The third model aimed to examine potential gender differences affecting the overall scores of anthropomorphism and animacy. Finally, we applied a likelihood ratio test to each model in order to test if the proportional odds assumption of the ordered logistic regression holds. This is a crucial assumption for ordinal regression analyses, and it asserts that the independent variables have the same effect on the odds irrespective of the splits between each pair of levels of the ordinal outcome variable. The null hypothesis of the test upholds the proportionality of the odds, thus for the assumption to be justifiable, it must not be rejected.  

\subsubsection{Attention and emotional content}
This analysis examines the effect of the CVAE conditioning variables \textit{v\_cond} (levels Negative, Neutral, Positive) and \textit{a\_cond} (levels Low, Medium, High) on the 5-point Likert scores of two items: \textit{Attention} (``The robot’s behaviour draws my attention'') and \textit{Emotion} (``The robot's expression was emotional''). For each of the Likert items, we fit an ordered logistic regression model and checked for pairwise differences with Tukey's adjustment of \textit{p} values. The proportional odds assumption is checked again with likelihood ratio tests. 

\subsection{Software}
For recording and executing robotic animations with a Pepper robot we used NAOqi 2.5 SDK\footnote{NAOqi 2.5:  \url{http://doc.aldebaran.com/2-5/home_pepper.html}}. For the robot simulations we used the Choregraphe Suite\footnote{Choregraphe Suite 2.5:  \url{http://doc.aldebaran.com/2-5/software/choregraphe/index.html}}. The code for the CVAE model was adapted from the TFModelLib collection of neural networks\footnote{TFModelLib repository:  \url{https://github.com/nhemion/tfmodellib}}. For the implementation of B-splines we used Splipy\footnote{Splipy v1.3.1: \url{https://sintefmath.github.io/Splipy/}}. The rest of the code for the CVAE implementation (data preprocessing, sampling, torus grids, etc.) was written in Python 3 with packages such as NumPy \cite{NumPy}, SciPy \cite{SciPy}, pandas \cite{pandas}, scikit-learn \cite{sklearn}, Matplotlib \cite{Matplotlib}. The interface for the user study data collection was written in Python 2 with Django v1.11.10\footnote{Django web framework: \url{https://www.djangoproject.com/}}. The statistical analysis of the collected data was carried out in R \cite{R, ggplot2, rstatix, MASS, ordinal, emmeans, dplyr, likert}. 

\section{Results}

\subsection{Valence, arousal and dominance ratings}
In Table \ref{table:vad_summary}, we present the descriptive statistics of the aggregated ratings (valence, arousal, and dominance), with respect to the levels of the explanatory variables \textit{v\_cond} and \textit{a\_cond}. 

\begin{table}
\caption{Summary statistics for valence, arousal and dominance ratings}
\centering
\label{table:vad_summary}
\begin{tabular}{l l l l l l l}
\toprule
Aggregated ratings & v\_cond level & Mean & SD & a\_cond level & Mean & SD\\
\midrule
\multirow{3}{*}{valence}  & Negative &  0.46 & 0.12 & Low & 0.47 & 0.1\\
  & Neutral &  0.54  & 0.09 & Medium & 0.54 & 0.11\\
  & Positive & 0.57 & 0.12 & High & 0.55 & 0.13\\
\hline
\multirow{3}{*}{arousal} & Negative &  0.55 & 0.12 & Low & 0.5 & 0.12 \\
  & Neutral & 0.55 & 0.12 & Medium & 0.55 & 0.12\\
  & Positive & 0.58  & 0.11 &   High & 0.63 & 0.14\\
\hline
\multirow{3}{*}{dominance}  & Negative & 0.4  & 0.12 & Low & 0.42 & 0.11\\
  & Neutral & 0.46  & 0.1   & Medium & 0.44 & 0.09\\
  & Positive & 0.52  &  0.13 &  High &  0.52 & 0.12\\
\bottomrule
\end{tabular}
\end{table}

Regarding the normality assumption, only two extreme outliers were detected, one in the relationship where \textit{v\_cond} is used to predict arousal, and a second one where \textit{a\_cond} is used to predict valence. Since these relationships are not substantial for our main hypotheses and we only have an exploratory interest in them, we decided to proceed. All three dependent variables---valence, arousal and dominance---were found normally distributed at each level of \textit{v\_cond} as assessed by Shapiro-Wilk’s test ($p > 0.05$). The same was true for arousal and dominance with respect to the levels of \textit{a\_cond}, but valence was found to violate the assumption for the \textit{a\_cond} level \textit{medium} ($p=0.029$). Again, taking into account that it is not in our main hypothesis to predict valence ratings with the arousal conditioning of the CVAE, this violation of the normality assumption was not of concern. Furthermore, one-way ANOVA is considered a robust test against the normality assumption.

First, we will discuss the results for the tests in which we used the valence conditioning of the CVAE (\textit{v\_cond}), to predict valence, arousal and dominance ratings. The results revealed a significant effect of \textit{v\_cond} on the aggregated valence ratings ($F(2, 38) = 13.5,~p < 0.001,~ \eta_{g}^{2} = 0.16$) and on the aggregated dominance ratings ($F(2, 38) = 12.6,~ p < 0.001,~ \eta_{g}^{2} = 0.16$). No significant effect was detected on arousal ($F(2, 38) = 1,~ p =0.38,~ \eta_{g}^{2} = 0.01$).

With regard to the arousal conditioning of the CVAE (\textit{a\_cond}), the one-way repeated-measures ANOVAs detected significant effects on the aggregated valence ratings ($F(2, 38) = 6.47,~p =0.004,~ \eta_{g}^{2} = 0.09$), the aggregated arousal ratings ($F(1.5, 28.42) = 10.19,~ p = 0.001,~ \eta_{g}^{2} = 0.15$, with Greenhouse-Geisser correction for the degrees of freedom due to sphericity assumption violation), and the aggregated dominance ratings ($F(2, 38) = 16.159,~ p <0.0001,~ \eta_{g}^{2} = 0.166$). 

% The results are summarized in Table \ref{table:CVAE_anovas_a_cond}, along with the violations of assumption (VoA).  

\begin{table}
\caption{Post hoc tests for valence, arousal and dominance ratings}
\centering
\label{table:CVAE_posthoc}
\begin{tabular}{l l l l l l l}
\toprule
DV & IV & Group 1 & Group 2 & t(df) & p & p (adjusted) \\
\midrule
\multirow{3}{*}{valence}  & \multirow{3}{*}{v\_cond} & Negative & Neutral & $t(19)=-3.65$ & 0.002 & $0.005^{**}$ \\
  & & Negative & Positive & $t(19)=-4.47$ & $<0.001$ & $<0.001^{***}$ \\
  & & Neutral & Positive & $t(19)=-1.63$ & 0.12     & $0.36$ \\
\hline
\multirow{3}{*}{dominance}  & \multirow{3}{*}{v\_cond} & Negative & Neutral & $t(19)=-2.06$ & 0.05 & $0.16$ \\
  & & Negative & Positive & $t(19)=-4.69$ & $<0.001$ & $<0.001^{***}$ \\
  & & Neutral & Positive & $t(19)=-3.46$ & 0.003     & $0.008^{**}$ \\
\hline
\multirow{3}{*}{arousal}  & \multirow{3}{*}{a\_cond} & Low & Medium & $t(19)=-2.27$ & 0.04 & $0.11$ \\
  & & Low & High & $t(19)=-3.62$ & 0.002 & $0.005^{**}$ \\
  & & Medium & High & $t(19)=-2.88$ & 0.01    & $0.029^{*}$ \\
\hline
\multirow{3}{*}{dominance}  & \multirow{3}{*}{a\_cond} & Low & Medium & $t(19)=-0.89$ & 0.381 & $1$ \\
  & & Low & High & $t(19)=-4.37$ & $<0.001$ & $<0.001^{***}$ \\
  & & Medium & High & $t(19)=-5.24$ & $<0.001$ & $<0.001^{***}$ \\
\bottomrule
\end{tabular}
\vspace{1ex}

   {\raggedright \textit{\small Note: DV = Dependent Variable, IV = Independent Variable, t(df) = t statistic and degrees of freedom, p = \textit{p} values, p (adjusted) = \textit{p} values with Bonferroni correction.} \par} 
\end{table}

Pairwise differences between the levels of conditioning were examined with post hoc tests and are summarized in Table \ref{table:CVAE_posthoc}. We excluded the relationships for which ANOVAs did not reveal significant effects (arousal with \textit{v\_cond} as a predictor), and also those for which the normality assumption was violated (valence with \textit{a\_cond} as a predictor).  

\begin{figure}[htbp!] 
  \centering
    \includegraphics[width=0.7\textwidth]{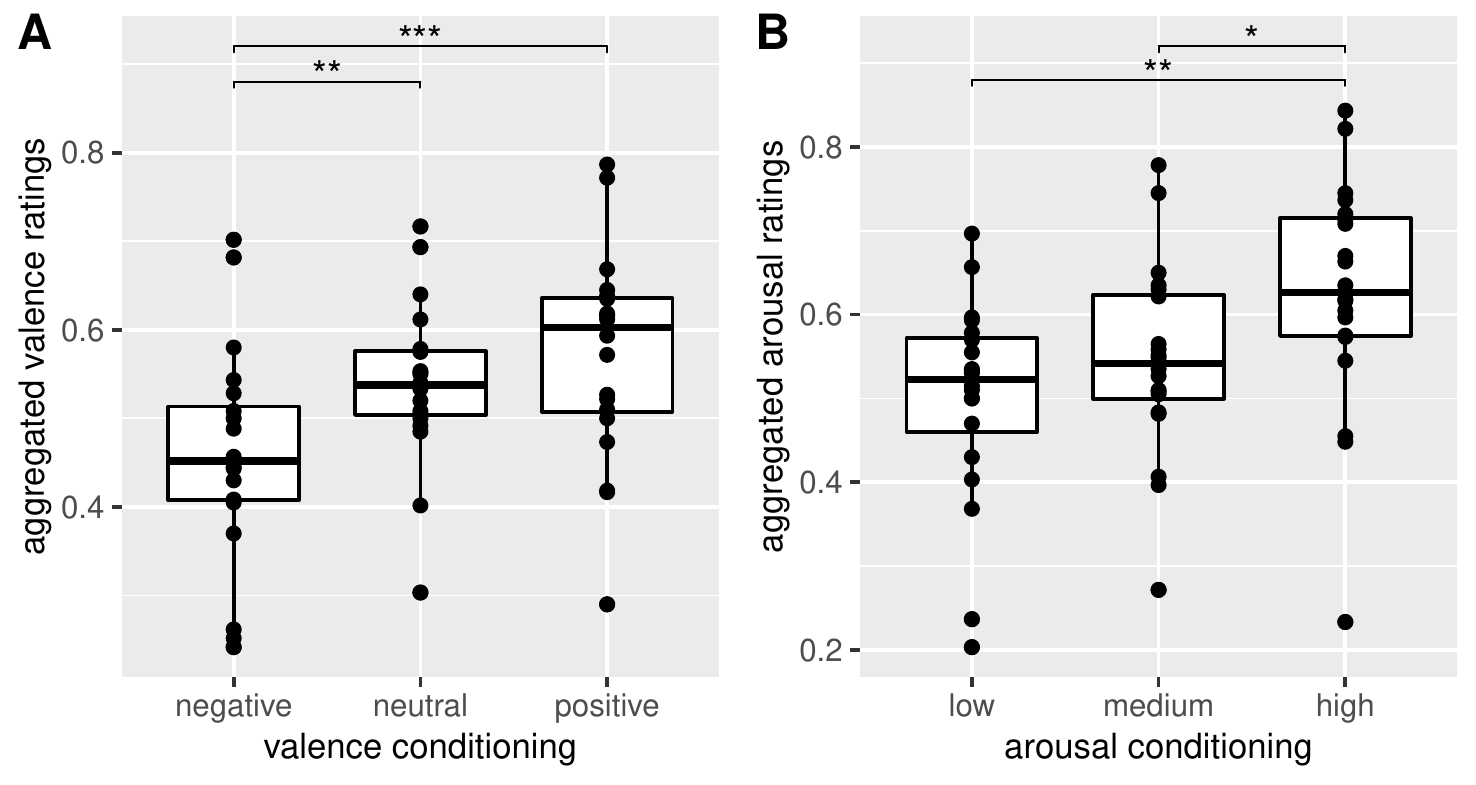}
    \caption[Pairwise test for valence and arousal conditioning levels with respect to the valence and arousal ratings]{Post hoc t-tests with Bonferroni adjustment. A) Valence conditioning impact on valence ratings. B) Arousal conditioning impact on arousal ratings. The graph was originally published in \cite{Marmpena2020}}.
    \label{fig:v_vcond_a_acond}
\end{figure}

For the effect of \textit{v\_cond} on valence ratings (Fig.~\ref{fig:v_vcond_a_acond}A), paired samples t-tests with Bonferroni correction revealed statistically significant differences for Negative-Neutral ($t=-3.65$,~ $p = 0.005$) and Negative-Positive ($t=-4.47$,~ $p <0.001$). Thus, the results partially support the hypothesis \textbf{H1}, in that animations generated with negative valence conditioning received lower valence ratings compared to animations generated with neutral or positive valence conditioning. 

For the effect of \textit{a\_cond} on the arousal ratings (Fig.~\ref{fig:v_vcond_a_acond}B), paired samples t-tests with Bonferroni correction detected statistically significant differences for Low-High ($t=-3.62$,~  $p = 0.005$) and Medium-High ($t=-2.88$,~ $p =0.029$). For Low-Medium a statistically significant difference was detected ($t=-2.27$,~ $p = 0.01$), but it did not survive the Bonferroni adjustment. Thus, the results partially support the hypothesis \textbf{H2}, in that animations generated with high arousal conditioning received higher arousal ratings compared to animations generated with low or medium arousal conditioning. 

\begin{figure}
  \centering
    \includegraphics[width=.7\textwidth]{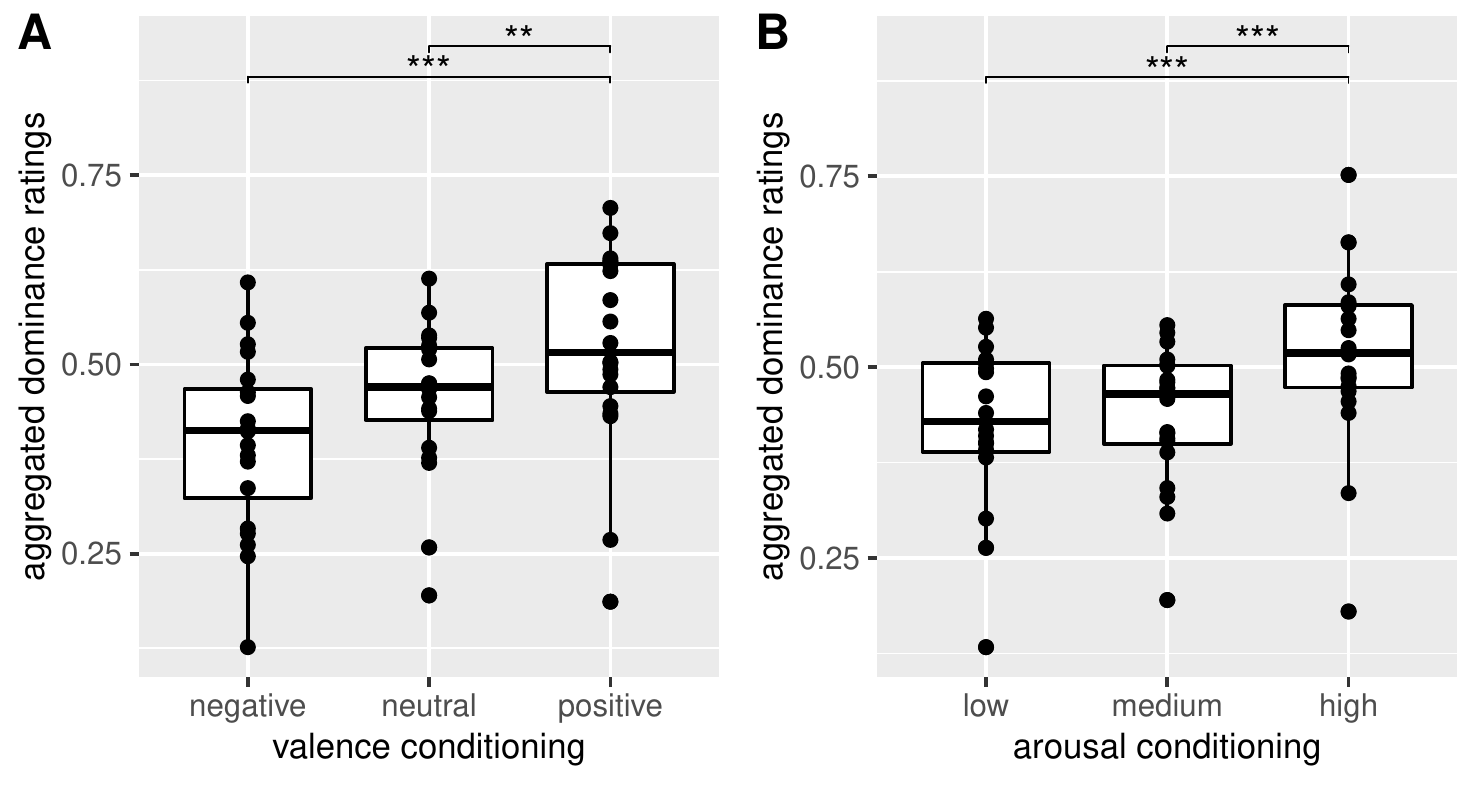}
    \caption[Pairwise test for valence and arousal conditioning levels with respect to the dominance ratings]{Post hoc t-tests with Bonferroni adjustment for multiple comparisons. A) Differences among the valence conditioning levels with respect to the participants' aggregated ratings of dominance. B) Differences among the arousal conditioning levels with respect to the participants' aggregated ratings of dominance.}
    \label{fig:d_vcond_d_acond}
\end{figure}

Furthermore, we conducted post hoc tests to examine the pairwise differences in the levels of \textit{v\_cond} and \textit{a\_cond} regarding their effect on the aggregated dominance ratings. For \textit{v\_cond}, statistically significant differences were detected for Negative-Positive ($t=-4.69$,~ $p < 0.001$) and Neutral-Positive ($t=-3.46$,~ $p = 0.008$)  (Fig.~\ref{fig:d_vcond_d_acond}A). For \textit{a\_cond}, statistically significant differences were detected for Low-High ($t=-4.37$,~ $p < 0.001$) and Medium-High ($t=-5.24$,~ $p < 0.001$)  (Fig.~\ref{fig:d_vcond_d_acond}B). These results suggest that the CVAE conditioning with \textit{v\_cond} and \textit{a\_cond} can potentially influence how people perceive the level of dominance in the robot's expression, with positive valence or high arousal conditioning making the robot to appear more dominant in terms of the simulated emotion.

\subsection{Comparison of designed and generated animations}
\begin{figure}[hb!] 
  \centering
    \includegraphics[width=\textwidth]{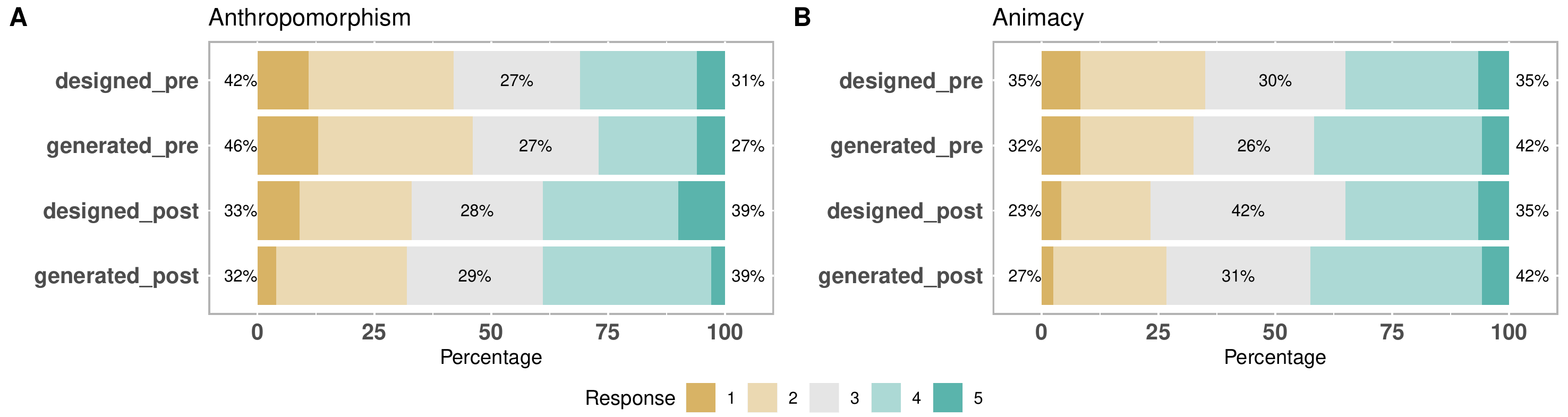}
    \caption[Bar plots for Anthropomorphism and Animacy scores]{Anthropomorphism (A) and Animacy (B) bar plots for designed vs generated animations in pretest and posttest phase. }
    \label{fig:anth_anim_bar_lik}
\end{figure}

We begin the presentation of the results with the descriptive statistics presented as bar plots in Fig.~\ref{fig:anth_anim_bar_lik} for the Anthropomorphism and Animacy scores. The responses are from 1 to 5 (with 5 indicating that the participant attributes a higher degree of Anthropomorphism or Animacy to the robot), and the percentages represent the portion of participants in each response. The basic contrast is \textit{designed} vs \textit{generated} animations presented in the pretest phase (Part A of the session) and then again in the posttest phase (Part C). 

For each scale (Anthropomorphism and Animacy), the internal consistency tested with Cronbach's alpha was found above our chosen threshold $\alpha \leq 0.85$. Thus, we proceeded with collapsing the semantic differentials of each scale (5 items for Anthropomorphism and 6 for Animacy) in one score per participant and per group, by taking the median value. 

\begin{figure}
  \centering
    \includegraphics[width=0.7\textwidth]{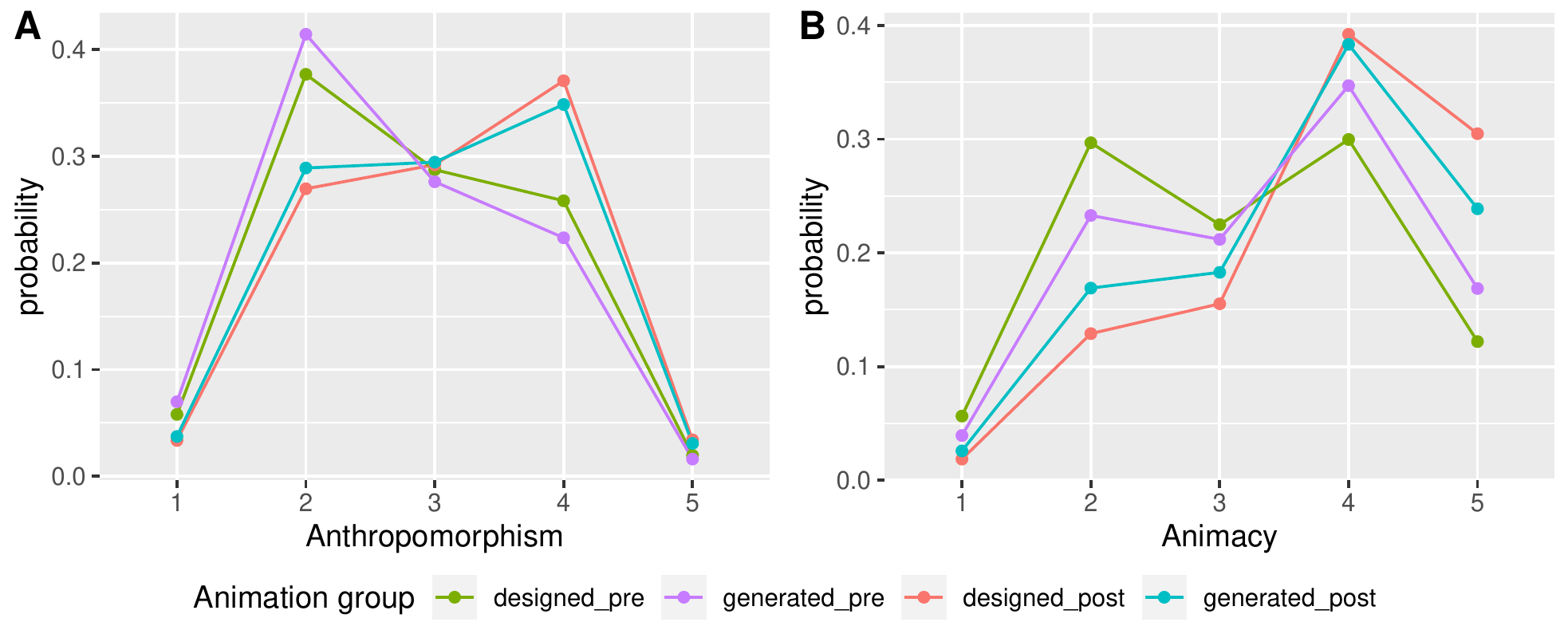}
    \caption[Anthropomorphism and Animacy scores for designed vs generated animations]{Anthropomorphism and Animacy scores for designed vs generated animations. The \textit{pre} and \textit{post} suffixes indicate the evaluation that was conducted during Part A and Part C of the experimental session respectively. }
    \label{fig:anth_anim_all}
\end{figure}

In terms of the hypothesis \textbf{H3} testing with ordered logistic regression, none of the groups (designed\_pre, generated\_pre, designed\_post, generated\_post) was found to have a statistically significant effect for Anthropomorphism or Animacy  (results are summarized in Table \ref{table:god_olr}). This result supports our hypothesis that participants do not attribute different Anthropomorphism or Animacy levels to the \textit{designed} animations compared to the \textit{generated} ones, either in the pretest or posttest phase. In Fig.~\ref{fig:anth_anim_all}, we plot the probabilities derived from the predictions of the two ordered logit models of Anthropomorphism and Animacy scores for each group of animations. 

\begin{table}[t!]
\caption{Ordered logistic regression results for Anthropomorphism and Animacy}
\centering
\label{table:god_olr}
\begin{tabular}{l l l l l l l }
\toprule
 DV & IV & Coef. & SE & t  & p  & 95\% CI\\ 
  \hline
\multirow{3}{*}{Anthropomorphism} & generated\_pre & -0.20 & 0.58 & -0.35 & 0.730 & (-1.34, 0.93)\\ 
  & designed\_post & 0.57 & 0.58 & 0.98 & 0.328 & (-0.57, 1.73)\\ 
  & generated\_post & 0.46 & 0.57 & 0.81 & 0.420 & (-0.66, 1.60)\\ 
  \hline                          
\multirow{3}{*}{Animacy} & generated\_pre & 0.38 & 0.67 & 0.57 & 0.572 & (-0.94, 1.71)\\ 
  & designed\_post & 1.15 & 0.66 & 1.75 & 0.080 & (-0.13, 2.46)\\ 
  & generated\_post & 0.81 & 0.69 & 1.18 & 0.240 & -0.54, 2.19)\\ 
  \hline
Anthropomorphism & posttest & 0.61 & 0.41 & 1.49 & 0.135 & (-0.19, 1.42)\\ 
\hline
Animacy  & posttest & 0.83 & 0.49 & 1.69 & 0.091 & (-0.12, 1.81)\\ 
\hline
Anthropomorphism & Male & -0.86 & 0.42 & -2.06 & $0.039^{*}$ & (-1.69, -0.05)\\ 
\hline
Animacy & Male & -0.45 & 0.49 & -0.92 & 0.359 & (-1.42, 0.5)\\ 
\hline
\end{tabular}
\vspace{1ex}

   {\raggedright \textit{\small Note: DV = Dependent Variable, IV = Independent Variable, Coef. = Coefficient of ordered logistic regression model, SE = Standard Error, t = t statistic, p = \textit{p} value,  CI = Confidence Interval.} \par} 
\end{table}

\begin{figure}
  \centering
    \includegraphics[width=0.7\textwidth]{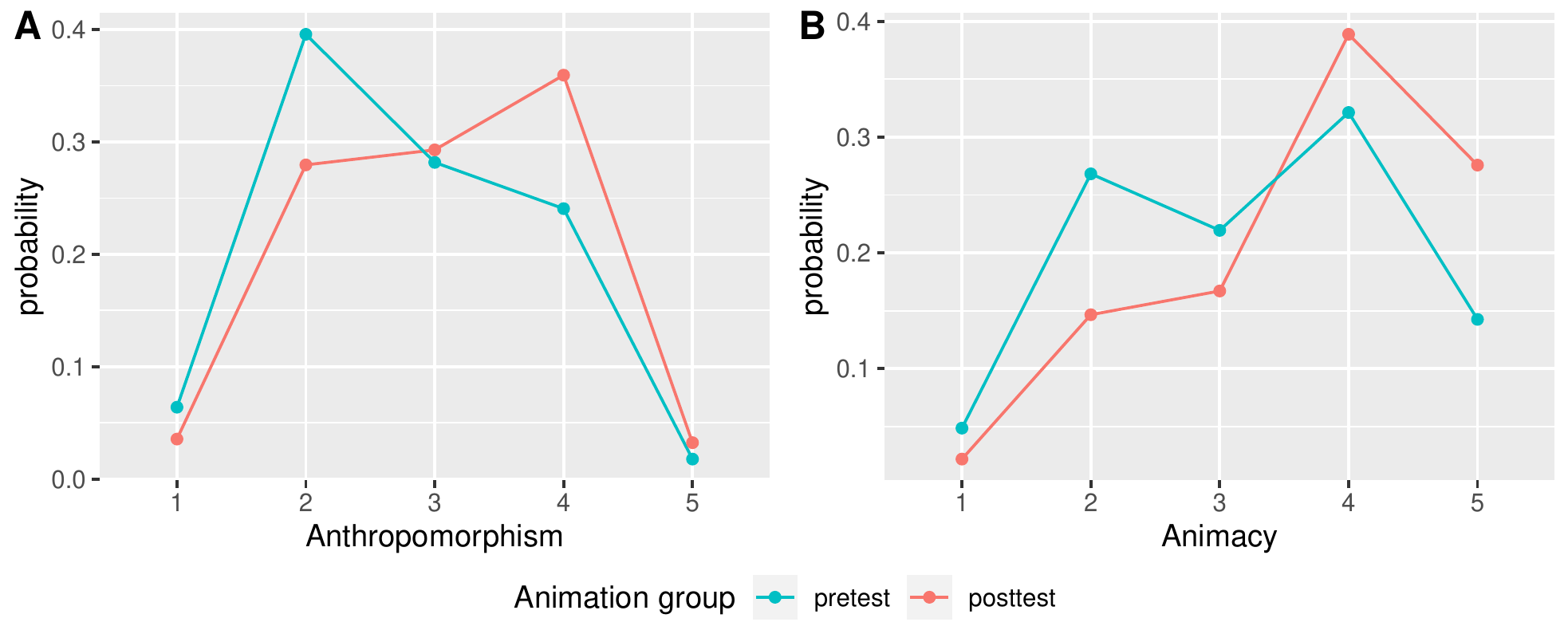}
    \caption[Anthropomorphism and Animacy scores in pretest vs posttest comparison]{Anthropomorphism and Animacy scores for all the animations evaluated in Part A and Part C. A slight increase in the posttest phase, is not confirmed with statistically significant results in pairwise comparisons.}
    \label{fig:anth_anim_pre_post}
\end{figure}

The two models of ordered logistic regression for Anthropomorphism and Animacy with all the pretest and all the posttest animations as a two-level predictor revealed some trends for Animacy (Fig.~\ref{fig:anth_anim_pre_post}B), but no significant effects were detected for either scale, suggesting that the interval in between the two evaluations did not impact significantly the perception of Anthropomorphism and Animacy. The detailed results are presented in Table \ref{table:god_olr}. 

\begin{figure}
  \centering
    \includegraphics[width=0.7\textwidth]{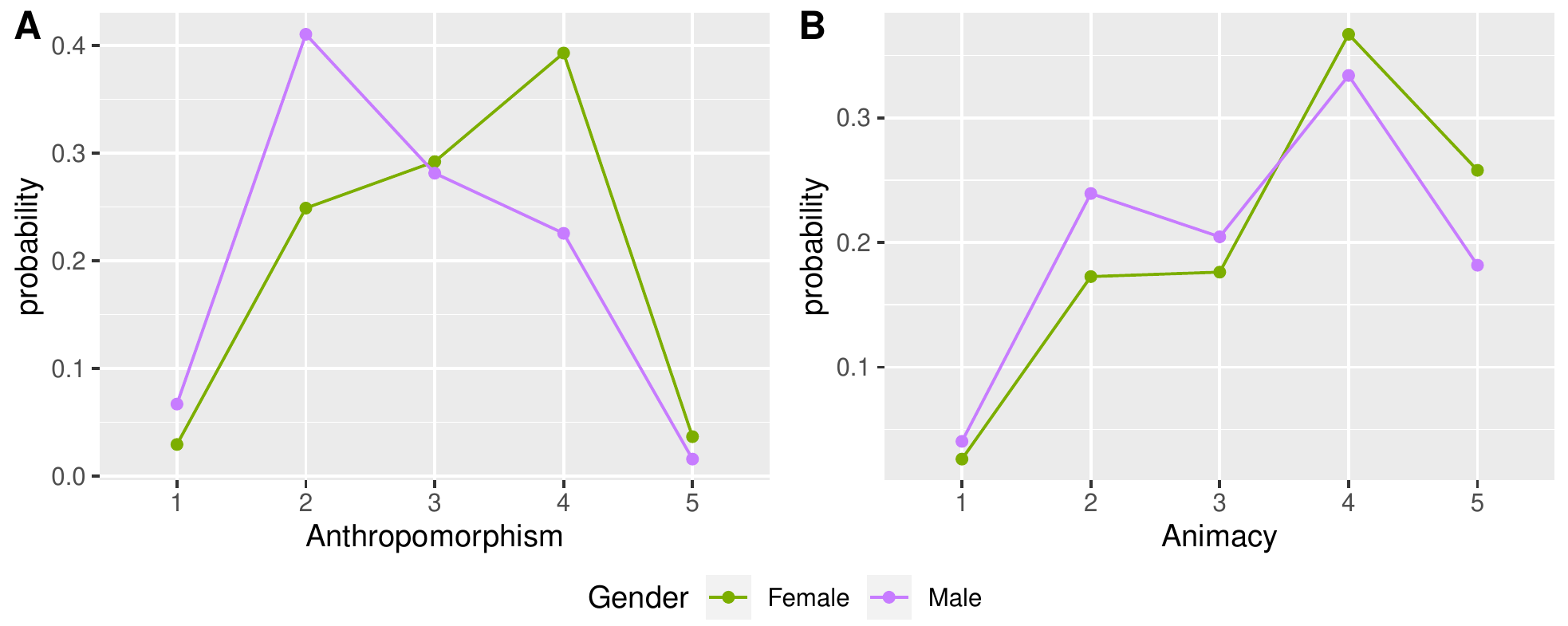}
    \caption[Anthropomorphism and Animacy scores in gender comparison]{Anthropomorphism and Animacy scores between male and female participants. Female participants gave significantly higher Anthropomorphism ratings.}
    \label{fig:anth_anim_sex}
\end{figure}

Finally, regarding the two models that tested for gender effects, female participants gave higher scores than male ($p = 0.039 $) on the Anthropomorphism scale, but no statistically significant differences were detected for Animacy ($p =  0.35$). The predictions' probabilities are plotted in Fig.~\ref{fig:anth_anim_sex} and the detailed results are included in Table \ref{table:god_olr}.

The likelihood ratio tests checking for violations of the proportional odds assumption did not obtain statistically significant \textit{p} values for any of the six models (all \textit{p} values $\leq 0.08$ ), thus the assumption was considered tenable. 

\subsection{Attention and emotional content}
\begin{figure}
  \centering
    \includegraphics[width=\textwidth]{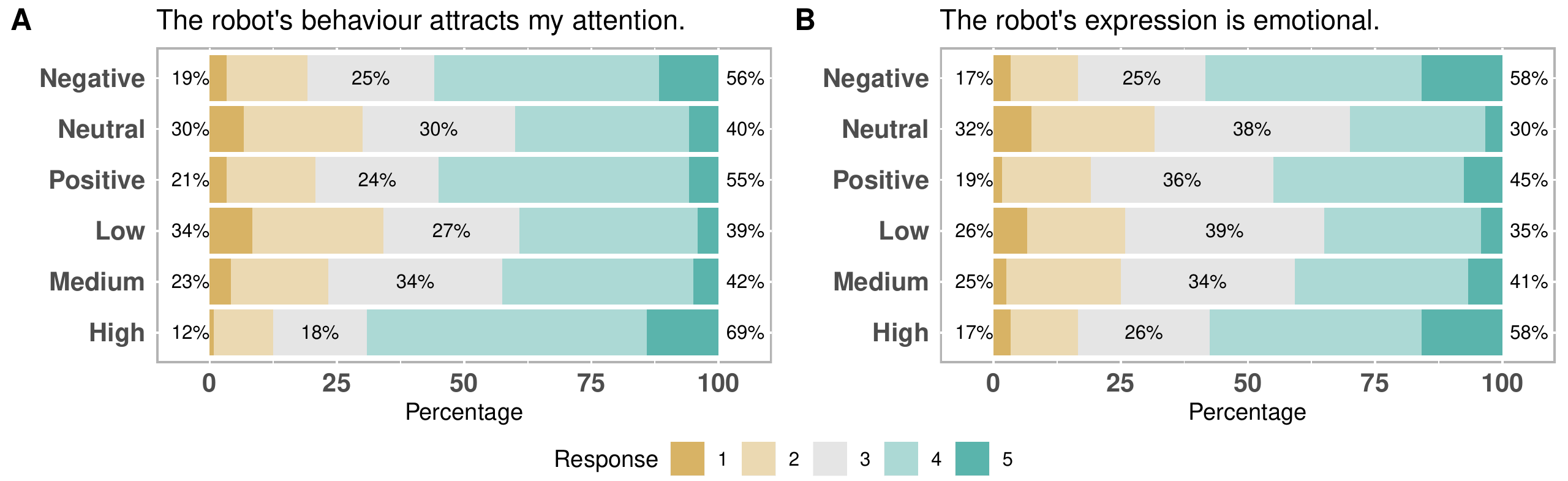}
    \caption[Bar plots for Attention and Emotion Likert scores]{Bar plots for Attention and Emotion Likert scores with respect to the valence (Negative, Neutral, Positive) and arousal (Low, Medium, High) conditioning levels.}
    \label{fig:att_emo_bar_lik}
\end{figure}

The descriptive statistics of the Attention and Emotion Likert scales with respect to the valence and arousal conditioning levels are presented in Fig.~\ref{fig:att_emo_bar_lik} as bar plots with the exact percentage for each response. The two ordered logistic regression models that predicted the Attention Likert scores (``The robot's behaviour draws my attention'') with the valence and arousal conditioning of the CVAE as a predictor respectively, detected statistically significant effects. The results are summarised in Table \ref{table:att_olr}. 

\begin{figure}[htbp!] 
  \centering
    \includegraphics[width=0.7\textwidth]{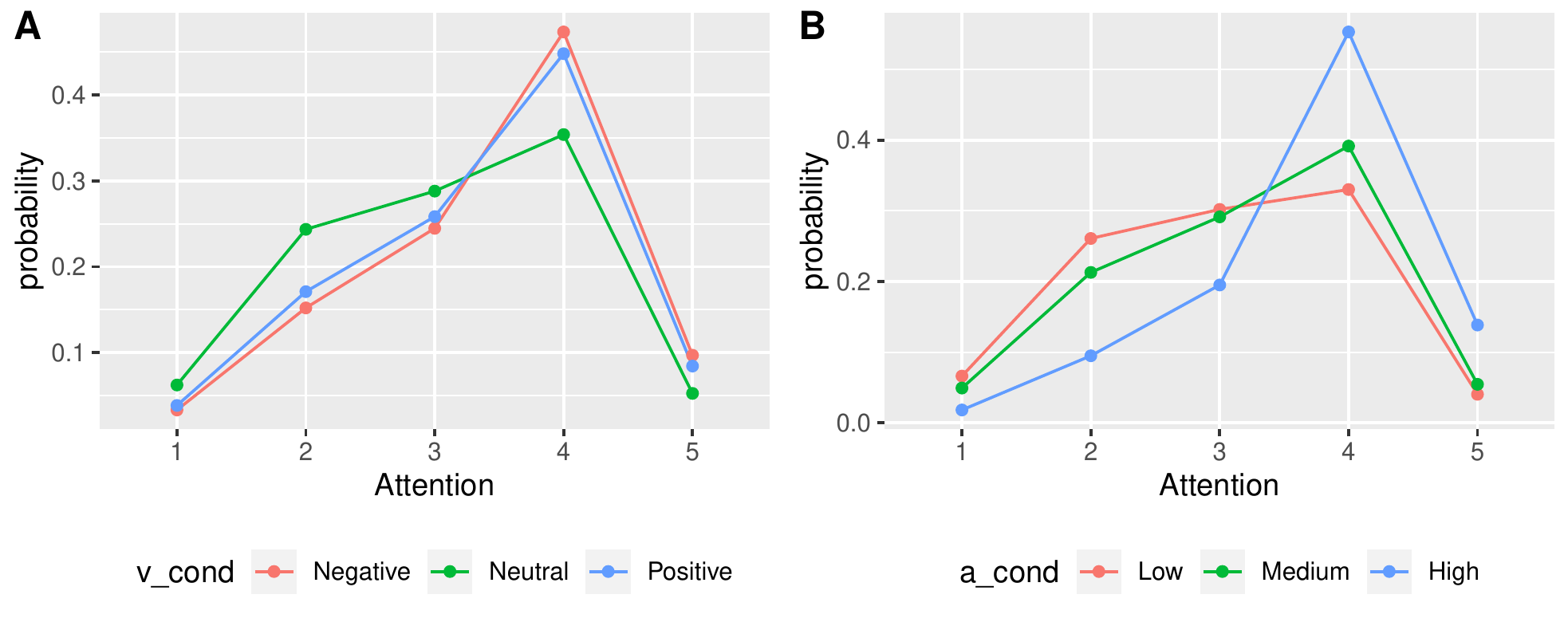}
    \caption[Attention scores for different levels of valence and arousal conditioning]{Attention scores (``The robot's behavior draws my attention'') for different levels of valence (A) and arousal (B) conditioning. A) Pairwise comparisons revealed significantly higher scores for negative conditioning compared to neutral. B) Pairwise comparisons revealed significantly higher scores for high arousal conditioning compared to both medium and low.}
    \label{fig:att_va}
\end{figure}

Post hoc tests on the levels of the valence conditioning (\textit{v\_cond}) determined that the contrast Negative-Neutral was statistically significant ($z=2.77 $,~ $p = 0.02$), suggesting that animations generated with negative valence conditioning might attract more attention from participants compared to animations generated with neutral valence. This result is illustrated in Fig.~\ref{fig:att_va}A, where we present the probabilities of predictions obtained by the model. In the same figure, we observe that the same appears to be true for animations generated with positive conditioning, however, this difference (Neutral-Positive) did not survive the Tukey adjustment of the \textit{p} value. The results are summarized in Table \ref{table:CVAE_posthoc_att}. 

\begin{table}[htbp!] 
\caption{Ordered logistic regression results for Attention and Emotion}
\centering
\label{table:att_olr}
\begin{tabular}{l l l l l l l }
\toprule
 DV & IV & Coef. & SE & t  & p  & 95\% CI\\ 
  \hline
\multirow{2}{*}{Attention}& v\_cond:Neutral & -0.11 & 0.17 & -0.64 & 0.52 & (-0.44, 0.22)\\ 
                          & v\_cond:Negative & 0.48 & 0.17 & 2.85 & $<0.001^{***}$ & (0.15, 0.81)\\
  \hline                          
\multirow{2}{*}{Attention}& a\_cond:Medium & 0.95 & 0.18 & 5.33 & $<0.001^{***}$ & (0.60, 1.30)\\ 
  & a\_cond:High & 0.29 & 0.17 & 1.73 & 0.08 & (-0.04, 0.62) \\ 
  \hline
\multirow{2}{*}{Emotion}&  v\_cond:Neutral & -0.36 & 0.17 & -2.10 & $0.04^{*}$ & (-0.69, -0.02) \\ 
  & v\_cond:Negative & 0.74 & 0.17 & 4.39 & $<0.001^{***}$ & (0.41, 1.07)\\ 
   \hline
\multirow{2}{*}{Emotion}&  a\_cond:Medium & 0.63 & 0.17 & 3.71 & $<0.001^{***}$ & (0.30, 0.97)\\ 
  & a\_cond:High & 0.19 & 0.17 & 1.18 & 0.24 & (-0.13, 0.52)\\ 
   \hline
\end{tabular}
\vspace{1ex}

   {\raggedright \textit{\small Note: DV = Dependent Variable, IV = Independent Variable, Coef. = Coefficient of ordered logistic regression model, SE = Standard Error, t = t statistic, p = \textit{p} value,  CI = Confidence Interval.} \par} 
\end{table}

\begin{table}[htbp!] 
\caption{Post hoc tests for Attention and Emotion}
\centering
\label{table:CVAE_posthoc_att}
\begin{tabular}{l l l l l l }
\toprule
DV & IV & Group 1 & Group 2 & z & p (adjusted) \\
\midrule
\multirow{3}{*}{Attention}  & \multirow{3}{*}{v\_cond} & Negative & Neutral & 2.77 & $0.02^{*}$ \\
  & & Negative & Positive & 0.64 &  $0.8$ \\
  & & Neutral & Positive & -2.16 & $0.08$ \\
\hline
\multirow{3}{*}{Attention}  & \multirow{3}{*}{a\_cond} & Low & Medium & -1.34  & 0.37 \\
  & & Low & High & -5.33 & $<.001^{***}$ \\
  & & Medium & High & -4.17 & $<.001^{***}$ \\
\hline
\multirow{3}{*}{Emotion}  & \multirow{3}{*}{v\_cond} & Negative & Neutral & 4.75 & $<.001^{***}$ \\
  & & Negative & Positive & 2.1 & $0.09$ \\
  & & Neutral & Positive & -2.81 & $0.01^{**}$ \\
\hline
\multirow{3}{*}{Emotion}  & \multirow{3}{*}{a\_cond} & Low & Medium & -0.89 & 0.64 \\
  & & Low & High & -3.71 & $<.001^{***}$ \\
  & & Medium & High & -2.85 & $0.01^{**}$ \\
\bottomrule
\end{tabular}
\vspace{1ex}

   {\raggedright \textit{\small Note: DV = Dependent Variable, IV = Independent Variable, z = z statistic, p (adjusted) = \textit{p} value with Tukey correction} \par} 
\end{table}

Post hoc tests with Tukey's \textit{p} value adjustment for the arousal conditioning (\textit{a\_cond}) revealed statistically significant differences for the contrasts Low-High ($z= -5.33$,~ $p<0.001$) and Medium-High ($z= -4.17$,~ $p<0.001$). The result suggests that potentially animations sampled with larger radius from the latent space tend to draw participants' attention more. The differences are presented in the model's predictions in Fig.~\ref{fig:att_va}B and the results are summarized in Table \ref{table:CVAE_posthoc_att}.

\begin{figure}[t!] 
  \centering
    \includegraphics[width=0.7\textwidth]{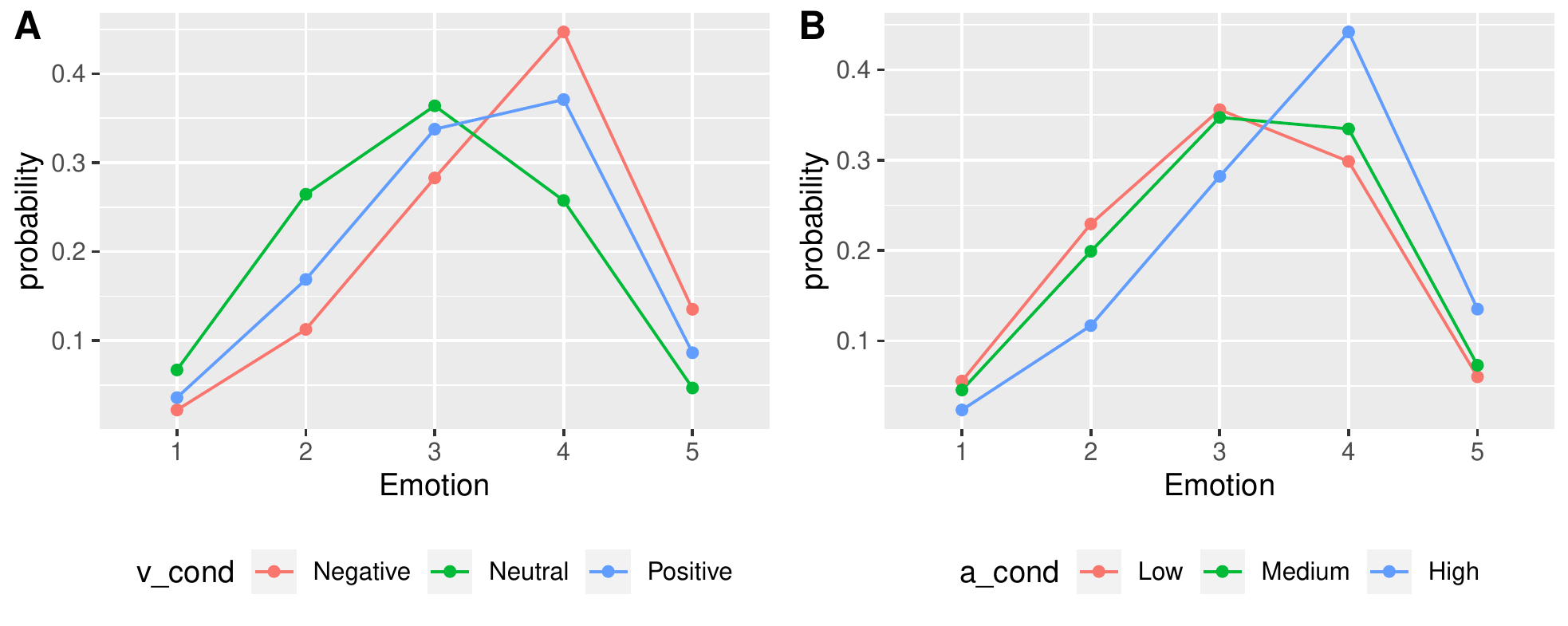}
    \caption[Emotion scores for different levels of valence and arousal conditioning]{Emotion scores (``The robot's expression is emotional'') for different levels of valence (A) and arousal (B) conditioning. A) Pairwise comparisons revealed significantly higher scores for negative and positive conditioning compared to neutral. B) Pairwise comparisons revealed significantly higher scores for high arousal conditioning compared to both medium and low.}
    \label{fig:emo_va}
\end{figure}

The two ordered logistic regression models predicting the Emotion Likert scores (``The robot's expression is emotional'') with the valence and arousal conditioning of the CVAE as a predictor respectively, detected statistically significant effects. The results are summarised in Table \ref{table:att_olr}. 
Post hoc tests with Tukey's \textit{p} value adjustment on the levels of the the valence conditioning (\textit{v\_cond}) obtained statistically significant results for the contrasts Negative-Neutral ($z= 4.75$,~  $p < 0.001$) and Neutral-Positive ($z= -2.81$,~ $p = 0.01$). The results imply (Fig.~\ref{fig:emo_va}A) that animations generated with negative or positive valence conditioning are perceived more as emotional compared to the ones generated with neutral valence conditioning. In the figure, a similar trend can perhaps be noted for Positive vs Negative, but this difference did not remain significant after the correction for multiple comparisons. The summary of the results is presented in Table \ref{table:CVAE_posthoc_att}.

For \textit{a\_cond} (arousal conditioning), statistically significant differences were detected for the contrasts Low-High ($z= -3.71 $,~ $p<0.001$) and Medium-High ($z= -2.85 $,~ $p=0.01$). This result suggests that animations sampled with a larger radius are perceived as more emotional. See also Fig.~\ref{fig:emo_va}B, and Table\ref{table:CVAE_posthoc_att} for a summary of the results. 

Regarding the proportional odds assumption, the likelihood ratio tests of the model terms produced \textit{p} values greater than 0.05 for all four models, thus the null hypothesis cannot be rejected and the assumption is tenable.

% The results of the tests are summarized in Table \ref{table:att_prop_asum}.

\textbf{}

\section{Conclusion}

In social robotics, human-robot interaction can be enhanced by endowing a robot with the ability to communicate using emotional body language (EBL). For robotic EBL to be engaging, it has to appear expressive, granular and interpretable in terms of the emotional class. To this aim, we implemented a Conditional Variational Autoencoder for the generation of multi-modal robotic EBL of targeted valence and arousal. The model was trained with a small set of robotic EBL animations designed for a Pepper robot by professional animators, including motion and eye LEDs colour sequences. The valence content of the generated animations was controlled by explicitly conditioning the training samples with scalar valence labels, while arousal conditioning was achieved by exploiting the geometry of the model's latent space to guide the sampling process. 

The interpretability of the generated animations was tested with a user evaluation study. The results provide support for valence conditioning which appears effective in differentiating positive or neutral from negative expressions. Similarly, arousal conditioning appears effective for differentiating high arousal from medium or low. However, only weak evidence were obtained for the differentiation between positive and neutral valence or between medium and low arousal. This limitation might be related to the composition of the training set, in which the representation of high valence and low arousal animations was lower than the rest of the classes \cite{Marmpena2018}. Such imbalance could suggest a sample bias within the dataset, i.e., the model was not trained with enough examples from these two categories and did not learn to differentiate them well. This is an issue which can be resolved by collecting more labelled animations of high valence and low arousal. Nevertheless, this is consistent with previous studies using a NAO and a Robovie \cite{Beck_interpretation_2010, Nakagawa2009} also report limitations in creating EBL expressions that are interpreted as of low arousal and high valence. Hence, it might be the case that this affect subspace is inherently difficult to represent, or perhaps emotional states from this affect subspace can be more successfully conveyed through other modalities, such as facial expression.

The analysis also showed that the generated animations do not appear to be less appealing compared to the designed animations in terms of Anthropomorphism and Animacy, either in the pretest phase or in the posttest phase. With regards to the impact of the robot's behaviour on the user's attention (presumably an aspect of believability), the results demonstrate that animations of more extreme levels of valence, or of high arousal, seem to draw more attention. Similarly, these animations are more strongly classified as emotional. A video with the physical robot executing several CVAE generated animations is available online\footnote{CVAE generated animation set video: \url{https://youtu.be/wmLT8FARSk0}}. The animations used for training, as well as the generated ones presented in the user study  in this study can also be found online\footnote{REBL-Pepper Dataset: \url{https://github.com/minamar/rebl-pepper-data}}.

\section*{Acknowledgments}
This project received funding from the European Union’s Horizon 2020 research and innovation program under the Marie Skłodowska Curie grant agreement No 674868 (APRIL).

\bibliographystyle{unsrt}  
\bibliography{references}

\end{document}